\documentclass[sigconf]{acmart}
\AtBeginDocument{%
  }

\usepackage{algorithm}
\usepackage{algorithmic}
\usepackage[capitalize,noabbrev]{cleveref}

\usepackage{multirow}
\usepackage{color}
\usepackage{xcolor}
\usepackage{colortbl}
\usepackage{tabularx}
\usepackage{arydshln}
\usepackage{bbm}
\usepackage[textsize=tiny]{todonotes}
\usepackage{dictsym}

\definecolor{Gray}{gray}{0.90}

\definecolor{mygray}{gray}{0.93}
\newcommand{\gt}[1]{\textcolor{gray}{#1}}

\theoremstyle{plain}
\newtheorem{theorem}{Theorem}[section]

\newtheorem{corollary}[theorem]{Corollary}
\theoremstyle{definition}

\newtheorem{assumption}[theorem]{Assumption}
\theoremstyle{remark}
\newtheorem{remark}[theorem]{Remark}

\copyrightyear{2025} 
\acmYear{2025} 
\setcopyright{acmcopyright}\acmConference[arXiv 2025]{arXiv}{2025}{arXiv}
\acmBooktitle{arXiv 2025}
\acmDOI{arXiv}
\acmISBN{arXiv}

\begin{document}

%%
%% The "title" command has an optional parameter,
%% allowing the author to define a "short title" to be used in page headers.
\title{BoxSeg: Quality-Aware and Peer-Assisted Learning for Box-supervised Instance Segmentation}

\settopmatter{authorsperrow=4}
\author{Jinxiang Lai *}
\affiliation{HKUST \country{Hongkong China}}

\author{Wenlong Wu *}
\affiliation{DJI \country{China}}

\author{Jiawei Zhan *}
\affiliation{Tencent \country{China}}

\author{Jian Li}
\affiliation{Tencent \country{China}}

\author{Bin-Bin Gao}
\affiliation{Tencent \country{China}}

\author{Jun Liu}
\affiliation{Tencent \country{China}}

\author{Jie ZHANG}
\affiliation{HKUST \country{Hongkong China}}

\author{Song Guo\dag}
\affiliation{HKUST \country{Hongkong China}}

\thanks{* Authors with same contributions}
\thanks{\dag Corresponding author}

%%
%% By default, the full list of authors will be used in the page
%% headers. Often, this list is too long, and will overlap
%% other information printed in the page headers. This command allows
%% the author to define a more concise list
%% of authors' names for this purpose.
\renewcommand{\shortauthors}{Jinxiang Lai, et al.}

%%
%% The abstract is a short summary of the work to be presented in the
%% article.
\begin{abstract}
Box-supervised instance segmentation methods aim to achieve instance segmentation with only box annotations.
Recent methods have demonstrated the effectiveness of acquiring high-quality pseudo masks under the teacher-student framework. 
Building upon this foundation, we propose a BoxSeg framework involving two novel and general modules named the Quality-Aware Module (QAM) and the Peer-assisted Copy-paste (PC).
The QAM obtains high-quality pseudo masks and better measures the mask quality to help reduce the effect of noisy masks, by leveraging the quality-aware multi-mask complementation mechanism.
The PC imitates Peer-Assisted Learning to further improve the quality of the low-quality masks with the guidance of the obtained high-quality pseudo masks.
Theoretical and experimental analyses demonstrate the proposed QAM and PC are effective.
Extensive experimental results show the superiority of our BoxSeg over the state-of-the-art methods, and illustrate the QAM and PC can be applied to improve other models.
\end{abstract}

%%
%% The code below is generated by the tool at http://dl.acm.org/ccs.cfm.
%% Please copy and paste the code instead of the example below.
%%


%%
%% Keywords. The author(s) should pick words that accurately describe
%% the work being presented. Separate the keywords with commas.
%% A "teaser" image appears between the author and affiliation
%% information and the body of the document, and typically spans the
%% page.
% \begin{teaserfigure}
%   \includegraphics[width=\textwidth]{sampleteaser}
%   \caption{Seattle Mariners at Spring Training, 2010.}
%   \Description{Enjoying the baseball game from the third-base
%   seats. Ichiro Suzuki preparing to bat.}
%   \label{fig:teaser}
% \end{teaserfigure}

% \received{20 February 2007}
% \received[revised]{12 March 2009}
% \received[accepted]{5 June 2009}

%%
%% This command processes the author and affiliation and title
%% information and builds the first part of the formatted document.
\maketitle

\section{Introduction}
Instance segmentation focuses on identifying and segmenting objects within images. Utilizing detailed mask annotations, instance segmentation techniques \cite{HeGDG17,YolactBolyaZXL19,CascdeCaiV21,SOLOV2WangZKLS20,SOLOWangKSJL20,TianSC20} have demonstrated remarkable performance on the challenging COCO dataset \cite{COCOLinMBHPRDZ14}. However, creating instance-level segmentation masks is significantly more complex and time-intensive compared to labeling bounding boxes. Recently, several studies \cite{BBTP,BoxInst,BBAM,BoxLevelSet,Boxcaseg,DiscoBox,li2023sim,cheng2023boxteacher} have investigated weakly-supervised instance segmentation using box annotations or color information. These weakly-supervised approaches can effectively train instance segmentation models \cite{HeGDG17,SOLOV2WangZKLS20,TianSC20} without the need for mask annotations, resulting in precise segmentation masks.

\begin{figure}[t]
\centering
\includegraphics[width=0.99\linewidth]{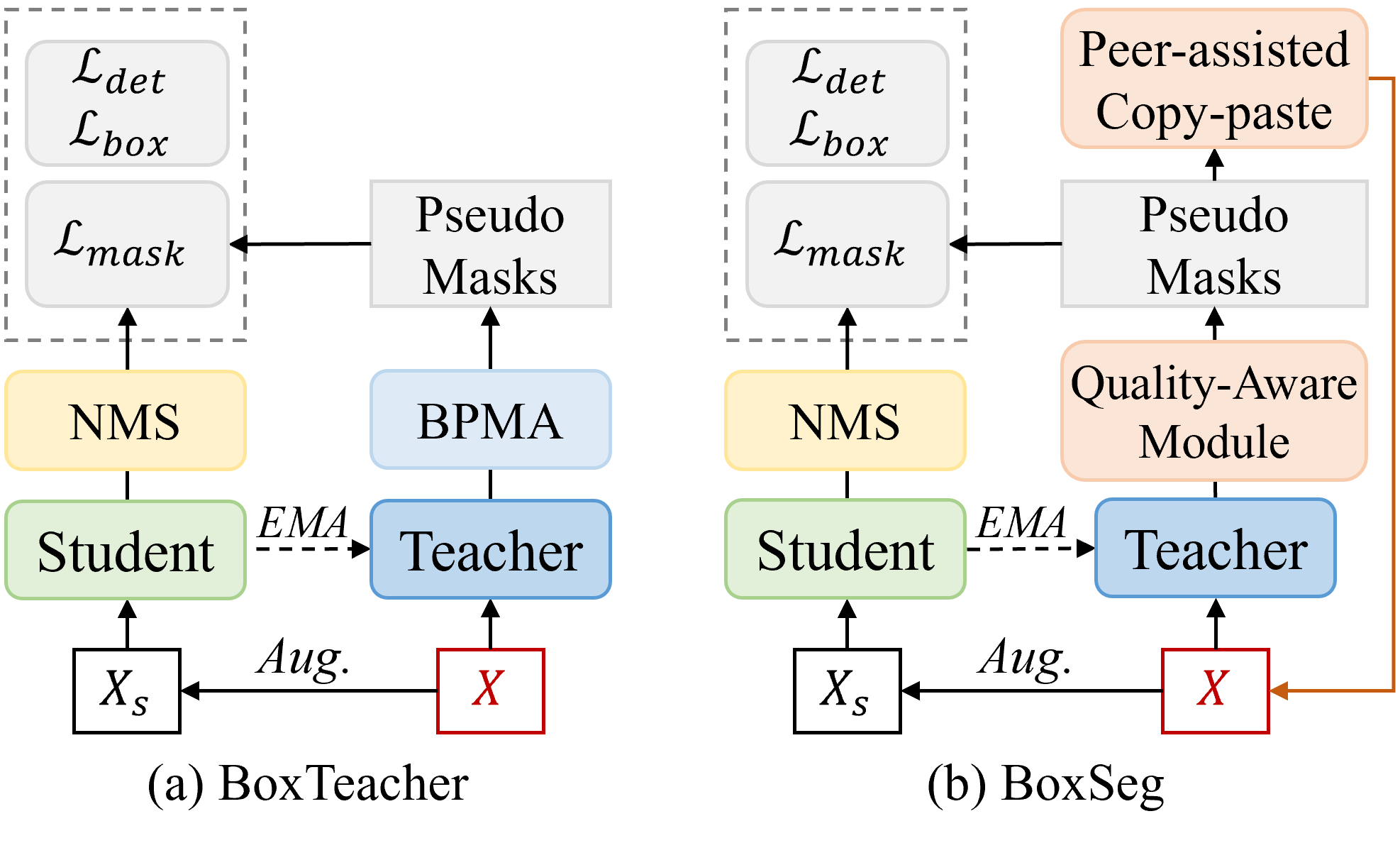}
\vspace{-5pt}
\caption{Compared to BoxTeacher, our BoxSeg integrates two novel modules named Quality-Aware Module and Peer-assisted Copy-paste to obtain high-quality pseudo masks and improve the quality of the pseudo masks respectively.}
\label{fig:framework}
\vspace{-1pt}
\end{figure}

Specifically, BoxInst \cite{BoxInst} can achieve instance segmentation with only box annotations, by replacing the original pixel-wise mask loss with the projection and pairwise affinity mask loss.
Since Box-Supervised Instance Segmentation (BSIS) approaches can predict some precise segmentation masks, BoxTeacher \cite{cheng2023boxteacher} and SIM \cite{li2023sim} proposed to optimize a student model with pseudo masks generated by a teacher model.
SIM implemented prototype-based segmentation to produce semantic-level pseudo masks.
BoxTeacher, as shown in Fig.\ref{fig:framework} (a), presented a Box-based Pseudo Mask Assignment (BPMA) to select high-quality pseudo masks, which is more effective than SIM.
Both BoxTeacher and SIM have demonstrated the effectiveness of acquiring high-quality pseudo masks under the teacher-student framework.

\begin{figure}[t]
\centering
\includegraphics[width=0.99\linewidth]{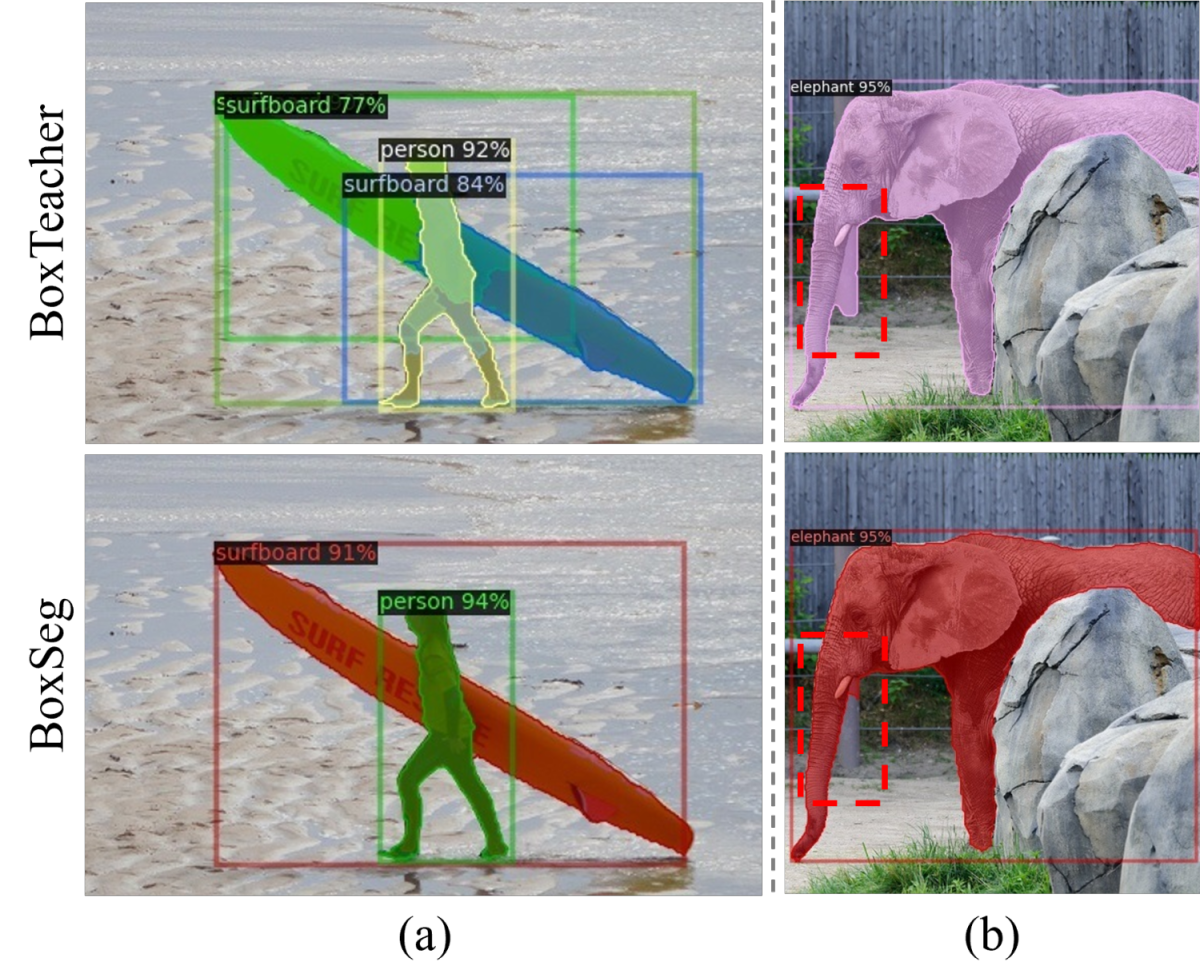}
\vspace{-5pt}
\caption{BoxTeacher struggles to (a) overlapped objects distraction and (b) similar background distraction, while BoxSeg can obtain more accurate masks under these distractions.}
\label{fig:motivation_vis}
\vspace{-1pt}
\end{figure}

However, as shown in Fig.\ref{fig:motivation_vis}, BoxTeacher produces redundant results under overlapped objects distraction and struggles to distinguish similar backgrounds.
Through analysis, we found some problems in the BPMA of BoxTeacher when obtaining pseudo masks.
BoxTeacher is built upon an instance segmentation method named CondInst \cite{TianSC20} which estimates the mask based on the predicted bounding box of the instance object.
Selecting pseudo masks using the BPMA involves performing Non-Maximum Suppression (NMS) to obtain a predicted instance box for each Ground Truth (GT) box. If the Intersection-Over-Union (IOU) between this predicted box and the GT box is larger than a threshold, the corresponding predicted mask is considered a valid pseudo mask.
According to the process of the BPMA, there are some problems as follows:
\emph{Problem \ding{172}}, NMS may filter out high-quality pseudo masks or select low-quality pseudo masks. Because NMS selects the predicted box based on the cls-score (classification score of the detection head), but the cls-score cannot provide a comprehensive measure of the predicted mask's quality. 
As illustrated in Fig.\ref{fig:motivation_vis} (a), when a surfboard is occluded and divided into multiple parts, the incomplete parts of a surfboard are selected by BPMA as different instances, resulting in redundant results.
\emph{Problem \ding{173}}, the pseudo masks obtained after NMS may have inaccurate predictions, i.e., including additional parts as shown in Fig.\ref{fig:motivation_vis} (b).

To alleviate these problems above and make further improvements, as presented in Fig.\ref{fig:framework} (b), we propose a BoxSeg framework involving two novel modules called Quality-Aware Module and Peer-assisted Copy-paste, which aim to \textit{obtain high-quality pseudo masks} and \textit{improve the quality of the low-quality masks}, respectively.
As shown in Fig.\ref{fig:motivation_vis}, our BoxSeg obtains more accurate masks than BoxTeacher by alleviating the aforementioned problems.
In detail, the proposed Quality-Aware Module based on the
quality-aware multi-mask complementation mechanism, implements Box-Quality Ranking, Quality-aware Masks Fusion, and Mask-Quality Scoring, which ensures that high-quality pseudo masks are retained, obtain more accurate pseudo masks, and better measure the mask quality, respectively.
The Peer-assisted Copy-paste imitates Peer-Assisted Learning \cite{topping1998peer} to improve the quality of the low-quality masks with the guidance of the high-quality pseudo masks.

To address \emph{Problem \ding{172}}, the proposed Box-Quality Ranking ensures that high-quality pseudo masks are not filtered out, by selecting the top-${K}$ candidate boxes for each pseudo mask based on the box-IOU (IOU between the proposal box and the GT box).
Instead of relying on the cls-score (may have small box-IOU) as BPMA did, our Box-Quality Ranking utilizes box-IOU to select a set of candidate boxes (with large box-IOU).
As illustrated in Fig.\ref{fig:motivation_vis} (a), our method can filter out the incomplete parts of a surfboard by Box-Quality Ranking and predict a complete surfboard.

To alleviate \emph{Problem \ding{173}}, our Quality-aware Masks Fusion integrates the candidate pseudo masks selected by the Box-Quality Ranking to obtain more accurate pseudo masks.
Different from BoxTeacher following a one-to-one pattern (i.e., one mask is predicted based on one box), our Quality-aware Masks Fusion adopts a many-to-one mechanism (i.e., one mask is predicted based on multiple boxes), which fuses predictions from multiple boxes resulting in more accurate masks and better removal of similar backgrounds.

Furthermore, considering that the obtained high-quality pseudo mask may still be noisy, we propose Mask-Quality Scoring to estimate the mask-quality score representing the quality of the pseudo mask, which is vital in the subsequent processes including training with the quality-aware mask-supervised loss and Peer-assisted Copy-paste.
Specifically, the mask-quality scores are used as the weights for quality-aware mask-supervised loss, which dynamically adjusts the weights for pseudo masks based on their quality, to \textit{reduce the influence of noisy masks}. In Peer-assisted Copy-paste, the mask-quality scores are utilized to select the peer objects with the high-quality pseudo masks.

After obtaining the high-quality pseudo masks through the Quality-Aware Module, we further introduce the Peer-assisted Copy-paste inspired by Peer-Assisted Learning \cite{topping1998peer} to improve the quality of the low-quality masks.
In a Peer-Assisted Learning setting, students take turns acting as both learners and peer tutors. The peer tutors provide guidance to the learners, meanwhile, the peer tutors reinforce their understanding by teaching the learners.
As shown in Fig.\ref{fig:pacp}, the proposed Peer-assisted Copy-paste implements two steps to imitate Peer-Assisted Learning, including Selecting Peer Tutors and Teaching Learners.

In general, our main contributions are as follows:

$\bullet$ Based on the quality-aware multi-mask complementation mechanism, we propose a novel Quality-Aware Module to obtain high-quality pseudo masks and better measure the mask quality to help reduce the effect of noisy masks. Besides, it is a flexible module that can be integrated into other teacher-student frameworks for BSIS task. 

$\bullet$ We introduce peer-assisted learning to improve the quality of the low-quality masks, i.e., the proposed Peer-assisted Copy-paste first collects the peer objects with the high-quality pseudo masks, then utilizes the peer objects to assist the optimization of the low-quality masks.
It is a general module that can be applied to any BSIS framework, such as SIM (teacher-student framework) and BoxInst (single-model framework).

$\bullet$ Theoretical and experimental analyses of the proposed modules are provided to demonstrate their effectiveness.

$\bullet$ Integrating with the Quality-Aware Module and the Peer-assisted Copy-paste, we propose a BoxSeg framework for BSIS task.
We conduct extensive experiments to validate the effectiveness of the proposed modules, and demonstrate the superiority of our BoxSeg over state-of-the-art methods.

\section{Related Work}
\textbf{Instance Segmentation.}
Instance segmentation algorithms can be roughly divided into two-stage and single-stage methods.
Two-stage methods \cite{HeGDG17,MSRCNNHuangHGHW19,BMaskChengWH020,PointRendKirillovWHG20} use bounding boxes from object detectors and RoIAlign \cite{HeGDG17} to extract region-of-interest (RoI) features for object segmentation. 
Single-stage approaches \cite{YolactBolyaZXL19,PolarMaskXieSSWLLSL20,TianSC20,MEInstZhangTSYY20} typically utilize single-stage object detectors \cite{SSD,FCOSTianSCH19} to locate and identify objects, subsequently generating masks via object embeddings or dynamic convolution \cite{DynamicConvChenDLCYL20}, e.g., CondInst \cite{TianSC20}.
Recently, transformer-based approaches \cite{DETRCarionMSUKZ20,queryinst,maskformer,mask2former} have achieved good advancements in instance segmentation.

A relevant instance segmentation method Mask Scoring R-CNN \cite{huang2019mask}, integrated a learnable MaskIoU head (4 convolution layers) to predict the mask quality score, due to having ground truth (GT) mask as supervision.
Different from Mask Scoring R-CNN, the box-supervised instance segmentation task has no GT mask as supervision, which makes obtaining high-quality masks and estimating the mask quality very challenging.
To overcome these challenges with GT box only, our methodology is using an approximate metric box-IOU to represent the candidate mask quality, then obtaining high-quality pseudo mask based on the quality-aware multi-mask complementation mechanism, and finally reducing the noise effect of pseudo mask by estimating the mask quality.

\noindent\textbf{weakly-supervised Instance Segmentation.}
Given the high cost of annotating instance mask, weakly-supervised instance segmentation using image-level labels or bounding boxes has garnered considerable attention. Several methods \cite{PRM,AhnCK19,iam_zhu,ArunJK20} leverage image-level labels to create pseudo masks from activation maps. More recently, numerous box-supervised techniques \cite{BBTP,DiscoBox,BoxInst,BoxLevelSet,li2023sim,cheng2023boxteacher} have combined multiple instance learning (MIL) loss or pairwise relation loss from low-level features to achieve impressive results using box annotations. BoxInst \cite{BoxInst} builds upon a single-stage instance segmentation framework CondInst, and employs a pairwise loss to ensure that proximal pixels with similar colors share the same label.
BoxTeacher \cite{cheng2023boxteacher} and SIM \cite{li2023sim} inherit the box supervision from BoxInst \cite{BoxInst}, but BoxTeacher concentrates more on producing high-quality pseudo masks and reducing the impact of noisy masks, while SIM constructs a group of category-wise prototypes to identify foreground objects and assign them semantic-level pseudo labels.

Both BoxTeacher and SIM have demonstrated the effectiveness of acquiring high-quality pseudo masks under the teacher-student framework. 
Building upon this foundation, we propose a BoxSeg framework involving two novel modules called Quality-Aware Module and Peer-assisted Copy-paste, which aim to obtain high-quality pseudo masks and improve the quality of the low-quality masks, respectively.

\begin{figure*}[t]
\centering
\includegraphics[width=0.8\linewidth]{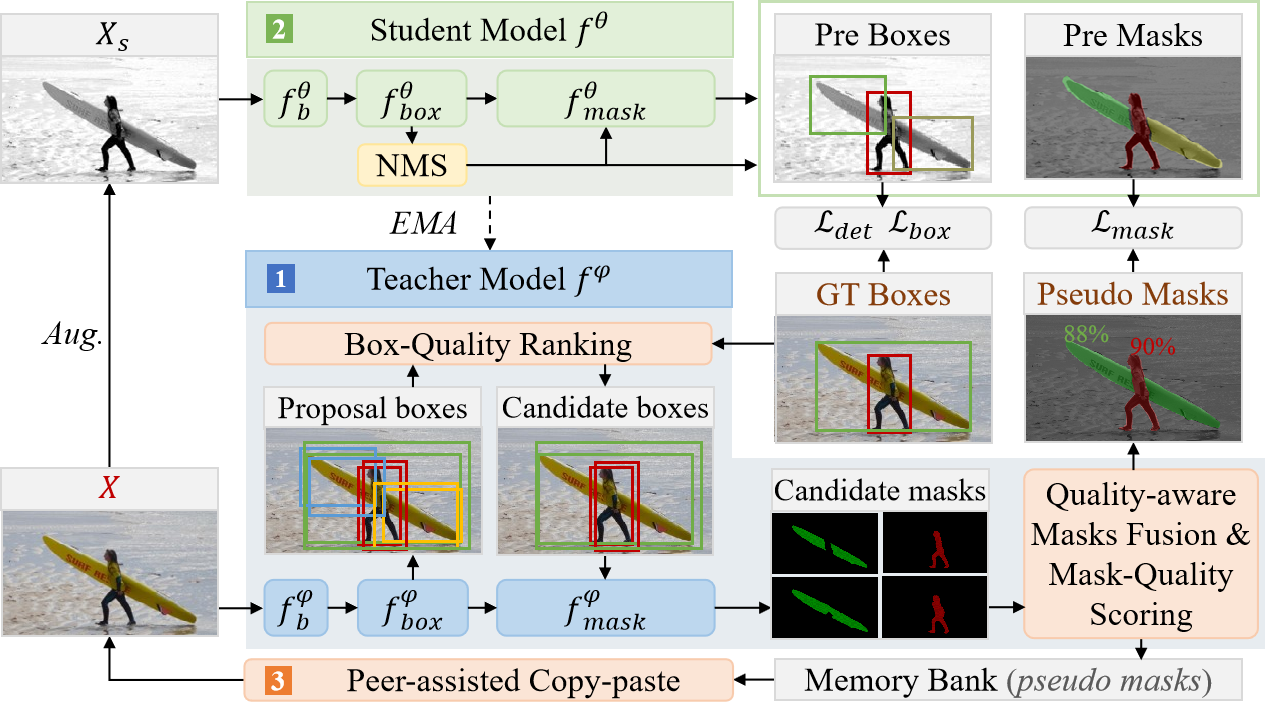}
\vspace{-5pt}
\caption{\textbf{The architecture of BoxSeg} consists of a Teacher Model, a Student Model, and Peer-assisted Copy-paste. \ding{172} The input image $X$ is processed by the \textit{Teacher Model} $f^{\varphi}$ with \textit{Quality-Aware Module} (i.e., including Box-Quality Ranking, Quality-aware Masks Fusion and Mask-Quality Scoring) to produce the pseudo masks and estimate their qualities. The CondInst is chosen as the basic segmentation network, which is composed of a backbone $f_{b}^{\varphi}$, a box branch $f_{box}^{\varphi}$, and a mask branch $f_{mask}^{\varphi}$. \ding{173} The \textit{Student Model} $f^{\theta}$ (i.e., CondInst model) predicts the image $X_s$ with augmentation and is supervised by the pseudo masks and the GT boxes, and then updates Teacher Model with exponential moving average (EMA). \textit{In the inference stage, the image is only processed by the Student Model to get the predictions of boxes and masks.} \ding{174} The \textit{Peer-assisted Copy-paste} copies and pastes the peer objects with high-quality pseudo masks into the image to assist the optimization of the low-quality masks.}
\label{fig:main_arch}
\vspace{-5pt}
\end{figure*}

\section{Methodology}
\subsection{Overview} 
In box-supervised instance segmentation task, only box annotated training data is provided, and is required to predict the bounding boxes and even the masks of the instance objects in the testing data.
Formally, box annotated data $\mathcal{D}=\{\mathcal{X}_i, \mathcal{Y}_i, \mathcal{B}_i\}^I_{i=1}$ is given for training, where $\mathcal{X}_i$ is the image, $\mathcal{Y}_i$ and $\mathcal{B}_i$ are a set of class and box labels in the $i$-th image, $I$ is the amount of images. 
In the testing stage, for each input image $\mathcal{X}_i$, the model needs to output the predictions $\{\mathcal{Y}_i, \mathcal{B}_i, \mathcal{M}_i\}$, where $\mathcal{M}_i$ is mask labels.

As shown in Fig.\ref{fig:framework} (b), our BoxSeg adopts the teacher-student framework, then integrates two novel modules named Quality-Aware Module and Peer-assisted Copy-paste to obtain high-quality pseudo masks and improve the quality of the pseudo masks respectively.
The proposed BoxSeg, an end-to-end training framework, integrates Teacher-Student Learning and Peer-Assisted Learning. In Teacher-Student Learning, the teacher model's prediction for each instance object is used as the pseudo mask to supervise the student model. Peer-Assisted Learning, on the other hand, involves using the peer object (with a high-quality pseudo mask) to assist the learning of the learner object (with a low-quality pseudo mask), fostering collaboration between two different instance objects.

\subsection{Architecture and Optimization}
\textbf{Architecture.}
The overall architecture of BoxSeg is presented in Fig.\ref{fig:main_arch}, which consists of a Teacher Model, a Student Model, and Peer-assisted Copy-paste.
The Teacher Model $f^{\varphi}$ and the Student Model $f^{\theta}$ adopt the same CondInst instance segmentation network with different learned parameters, differently, Teacher Model integrates Quality-Aware Module to obtain high-quality pseudo masks.

\noindent\textbf{Training Loss.}
The Student Model is end-to-end optimized under the supervision of GT box and pseudo mask generated by the Teacher Model, then the Teacher Model is updated with EMA from the Student Model, and the Peer-assisted Copy-paste is a non-parameter module.
Therefore, the loss of BoxSeg is: $\mathcal{L}\!=\!\mathcal{L}_{det}\!+\!\mathcal{L}_{box}\!+\!\mathcal{L}_{mask}$, which contains detection loss $\mathcal{L}_{det}$, box-supervised loss $\mathcal{L}_{box}$, and mask-supervised loss $\mathcal{L}_{mask}$.
The $\mathcal{L}_{det}$ and $\mathcal{L}_{box}$ are inherited from CondInst \cite{TianSC20} and BoxInst \cite{BoxInst}, respectively.
We present the \textit{quality-aware mask-supervised loss} $\mathcal{L}_{mask}$ defined as follows:
\begin{equation}
\label{pseudo_loss}
\mathcal{L}_{mask}\!=\! \frac{1}{N}\!\sum_{i=1}^{N}w_i\cdot\mathcal{L}_{\text{pixel}}(m_i^s,m_i^t),
\end{equation}
where $N$ denotes the number of valid teacher-generated pseudo masks, $w_i$ is the $i$-th mask-quality score estimated by Eq.~\eqref{eq:maskscore_definition} with our Quality-Aware Module, $m_i^s$ and $m_i^t$ denotes the $i$-th student-predicted mask and teacher-generated pseudo mask fused by Eq.~\eqref{eq:QMF}, $\mathcal{L}_{\text{pixel}}$ is pixel-wise segmentation loss like dice loss \cite{dice}.
In Eq.~\eqref{pseudo_loss}, the proposed mask-quality score $w_i$ adaptively scales the weight for pseudo mask loss, thus can take advantage of high-quality pseudo masks in a mask-supervised manner while reducing the influence of noisy masks, making $\mathcal{L}_{mask}$ be a quality-aware mask-supervised loss.
More analysis is presented in \cref{sec:ana_qml}.

\subsection{Quality-Aware Module}
The proposed Quality-Aware Module, consisting of Box-Quality Ranking, Quality-aware Masks Fusion, and Mask-Quality Scoring, which ensures that high-quality pseudo masks are retained, obtain more accurate pseudo masks, and better measure the mask quality, respectively.

\textbf{Box-Quality Ranking.} 
The proposed Box-Quality Ranking ensures that high-quality pseudo masks are not filtered out.
It selects the top-${K}$ candidate boxes $B_i=\{b_{i,n}\}^{K}_{n=1}$ for each pseudo mask $m_i^t$, based on the box-IOU values between the proposal boxes and the GT box.
Instead of relying on the cls-score, our Box-Quality Ranking utilizes box-IOU to select a set of candidate boxes.
The predicted box with a large box-IOU is the basic condition to predict a complete mask, thus box-IOU is a better basic measurement than cls-score to select the candidates containing the optimal pseudo mask.
Besides, the cls-score is unreliable in the early training stage, while the box-IOU is precise since the GT box is given.

\textbf{Quality-aware Masks Fusion.} 
To obtain more accurate pseudo masks, the Quality-aware Masks Fusion (QMF) integrates the candidate pseudo masks $M_i=\{m_{i,n}\}^{K}_{n=1}$ corresponding to the top-${K}$ candidate boxes $B_i=\{b_{i,n}\}^{K}_{n=1}$ selected by the Box-Quality Ranking (note that each predicted box $b_{i,n}$ has a corresponding predicted mask $m_{i,n}$ with the usage of CondInst).
The Quality-aware Masks Fusion integrates the candidate pseudo masks using the normalized box-quality score $\in [0,1]$ as weights which are the normalized values of the geometric mean of box-scores and box-IOUs, formally:
\begin{equation}
m_i^t=\sum_{n=1}^{K} {\frac{{ \mathbbm{1}(s_{i,n}>\tau_m)\cdot \sqrt{s_{i,n} \cdot u_{i,n}}}}
{\sum_{k=1}^{K} {\mathbbm{1}(s_{i,k}>\tau_m)\cdot \sqrt{s_{i,k} \cdot u_{i,k}}}} \cdot m_{i,n}},
\label{eq:QMF}
\end{equation}
where $s_{i,n}$ is the box-score (i.e., the IOU-aware classification score of the detection head as introduced in VarifocalNet \cite{zhang2021varifocalnet}) representing the quality of the predicted box, $u_{i,n}$ is the box-IOU between the predicted box and the GT box, $\mathbbm{1}(\cdot)$ is the indicator function, $\tau_m$ is the threshold.
We utilize the geometric mean of the box-score $s_{i,n}$ and box-IOU $u_{i,n}$ to represent the box-quality score $\sqrt{s_{i,n} \cdot u_{i,n}}$, since the box-IOU is accurate in training which can be used to rectify the box-score predicted by the model.
Then, the normalized box-quality score are used to approximate the mask quality for the weighted fusion of the candidate pseudo masks.
Thus, the contribution of each candidate mask to the final fused result is balanced, leading to a high-quality pseudo mask. 
We give the theoretical analysis for QMF, which provides formal guarantees on the generalization error of the fused mask $m_i^t$.

\begin{theorem}
The Upper Bound of the generalization error of the fused mask $m_i^t$ is as follows (proof is in \cref{sec:ana_qmf}):
\begin{equation}
\begin{aligned}
\text{Error}(m_i^t) &\leq \underbrace{\max_{n} \|m_{i,n} - m_i^*\|}_{\text{Approximation Error}} + \underbrace{O\left(\sqrt{\frac{\log(1/\delta)}{\hat{K}}}\right)}_{\text{Estimation Error}} \\
&+\underbrace{\epsilon_w \cdot \max_{n} \|m_{i,n}\|}_{\text{Weighting Error}},
\end{aligned}
\label{eq:error_bound}
\end{equation}
with probability at least $1 - \delta$, where $m_i^*$ is the true mask, $\hat{K} = \sum_{n=1}^{K} \mathbbm{1}(s_{i,n} > \tau_m)$, $\epsilon_w$ is the maximum error in the box-quality score.

\end{theorem}

\begin{remark}
Reducing the \textit{Approximation Error} through the integration of diverse candidate masks.
\end{remark}
\begin{remark}
\label{remark:qmf_k}
Controlling the \textit{Estimation Error} by increasing the number of effective candidate masks $\hat{K}$.
\end{remark}
\begin{remark}
Minimizing the \textit{Weighting Error} through the use of accurate box-quality score and optimal threshold $\tau_m$.
\end{remark}
\begin{remark}
\label{remark:qmf_tm}
A higher $\tau_m$ reduces the \textit{Weighting Error}, but increases the \textit{Estimation Error} due to the decrease of $\hat{K}$.
\end{remark}

However, in the original QMF defined by Eq.\ref{eq:QMF}, a fixed number of candidate masks $K$ may under-utilize large objects or over-smooth small ones.
To further improve pseudo-mask quality, we propose an adaptive-$K$ mechanism in QMF based on Eq.\ref{eq:error_bound}, where $K$ is adjusted based on object size. Formally,
\begin{equation}
K = 
\begin{cases} 
K_{\min}, & \text{if } a_i \leq A_s, \\ 
K_{\min} + (K_{\max}-K_{\min}) \cdot \dfrac{\sqrt{a_i} - \sqrt{A_s}}{\sqrt{A_l} - \sqrt{A_s}}, & \text{if } A_s < a_i < A_l, \\ 
K_{\max}, & \text{if } a_i \geq A_l 
\end{cases}
\label{eq:adaptive_k}
\end{equation}
where, $K_{\min}$ and $K_{\max}$ are lower and upper bounds for $K$, which are empirically set to $K_{\min}$ = 2 and $K_{\max}$ = 10. 
$A_l$ and $A_s$ are area thresholds for large and small objects, which are set to $A_l$ = $96^2$ and $A_s$ = $32^2$, following $AP_l$ and $AP_s$ conventions. 
$a_i$ is the area of the GT box for object $i$.
According to Eq.\ref{eq:error_bound}, adaptive-$K$ optimizes the error bound per object size.
For large objects ($K$ = $K_{\max}$), Estimation Error $\downarrow$ due to larger $\hat{K}$, and Approximation Error $\downarrow$ by integrating more diverse high-quality candidates.
For small objects ($K$ = $K_{\min}$), Weighting Error $\downarrow$ due to fewer noisy candidates, and Estimation Error $\uparrow$ but still controlled by $K \geq K_{\min}$.
For medium objects, $K$ scales linearly with $\sqrt{a_i}$.

\textbf{Mask-Quality Scoring.} 
Considering that the obtained high-quality pseudo mask may be still noisy, our Mask-Quality Scoring (MQS) estimates the mask-quality score to represent the quality of the pseudo mask.
The mask-quality score $w_i$ is defined as follows:
\begin{equation}
\begin{aligned}
\label{eq:maskscore_definition}
&w_i = \sqrt{\hat{s}_{i} \cdot \hat{m}_{i}^t}, \quad where,\\
&\hat{s}_{i} = \frac{\sum_{n=1}^{K}\mathbbm{1}(s_{i,n}>\tau_m)\cdot s_{i,n}}{\sum_{n=1}^{K}\mathbbm{1}(s_{i,n}>\tau_m)},\\
&\hat{m}_{i}^t = \frac{\sum^{H,W}_{x,y}\mathbbm{1}(m_{i,(x,y)}^t>\tau_m)\cdot m_{i,(x,y)}^t\cdot m^b_{i,(x,y)}}{\sum^{H,W}_{x,y}\mathbbm{1}(m_{i,(x,y)}^t>\tau_m)\cdot m^b_{i,(x,y)}},
\end{aligned}
\end{equation}
where $m^b_i \in \mathbbm{R}^{H\times W}$ is the binary mask of GT box corresponding to the pseudo mask $m_i^t \in \mathbbm{R}^{H\times W}$, and $m^b_i$ is used to filter out the regions of $m_i^t$ that are outside the GT box.
According to Eq.~\eqref{eq:maskscore_definition}, the mask-quality score $w_i$ takes into account the global box-score $s_{i,n}$ and the local pixel-wise mask probabilities $m_{i,(x,y)}^t$ to better measure the mask quality.
First, the higher average box-score $\hat{s}_{i}$ represents the pseudo mask $m_i^t$ is fused based on higher quality boxes, which is a basic measurement to approximate the global mask quality. 
Second, $\hat{m}_{i}^t$ is the average pixel-wise probability score of the mask inside GT box, and the higher score means more confident pixels in the mask.
We give the theoretical analysis for MQS, which provides formal guarantees on the estimation error of the estimated quality score $w_i$.

\begin{theorem}
For any $\epsilon > 0$, the estimation error of $w_i$ is bounded with high probability as follows (proof is in \cref{sec:ana_mqs}):
\begin{equation}
\mathbb{P}\left(|w_i - w_i^*| \geq \epsilon\right) \leq 2 \exp\left(-\frac{2\hat{K}\epsilon^2}{\sigma_s^2}\right) + 2 \exp\left(-\frac{2\hat{M}\epsilon^2}{\sigma_m^2}\right)
\end{equation}
where, the true mask quality is defined as $w_i^* = \text{IOU}(m_i^t, m_i^*)$.
$\{\mu_s,\mu_m\}$ and $\{\sigma_s^2,\sigma_m^2\}$ \{mean, variance\} of $\{\hat{s}_i,\hat{m}_i^t\}$.
$\hat{M} = \sum_{x,y}^{H,W} \mathbbm{1} (m_{i,(x,y)}^t > \tau_m)$.
\end{theorem}
\begin{corollary}
As $\hat{K} \to \infty$ and $\hat{M} \to \infty$, $w_i$ converges to $w_i^*$:
\begin{equation}
\lim_{\hat{K}, \hat{M} \to \infty} w_i = \sqrt{\mu_s \cdot \mu_m} \approx w_i^*.
\end{equation}
\end{corollary}

\begin{remark}
\label{remark:mqs_k}
Increasing $\hat{K}$ reduces the estimation error of $\hat{s}_i$, because the Hoeffding bound scales as $\exp\left(-\frac{2\hat{K}\epsilon^2}{\sigma_s^2}\right)$.
\end{remark}
\begin{remark}
\label{remark:mqs_m}
Increasing $\hat{M}$ reduces the estimation error of $\hat{m}_i^t$, because the Hoeffding bound scales as $\exp\left(-\frac{2\hat{M}\epsilon^2}{\sigma_m^2}\right)$.
\end{remark}
\begin{remark}
\label{remark:mqs_tm}
A lower $\tau_m$ increases $\hat{K}$ and $\hat{M}$, reducing the variance and tightening the error bound, but may include noisier samples in the estimates.
\end{remark}

\textbf{Discussion for Key Parameters.}
(1) $\tau_m$: According to \cref{remark:qmf_tm} and \cref{remark:mqs_tm}, $\tau_m$ introduces a trade-off with $\hat{K}$ and $\hat{M}$. We empirically set $\tau_m=0.5$.
(2) $\hat{K}$: According to \cref{remark:qmf_k} and \cref{remark:mqs_k}, increasing $\hat{K}$ leads to better results for large objects. In practice, our experiments show that the performances are stable when $K_{\max}$ is larger than 10. Besides, larger $K_{\max}$ will increase the computation cost. Thus, we set $K_{\max}$ to 10.
(3) $\hat{M}$: Tab.\ref{tab:main_experiments} show that our BoxSeg achieves remarkable results on
large objects according to \textbf{AP$_{l}$} metric, which is consistent with \cref{remark:mqs_m}.

\begin{figure}[t]
\centering
\includegraphics[width=0.85\linewidth]{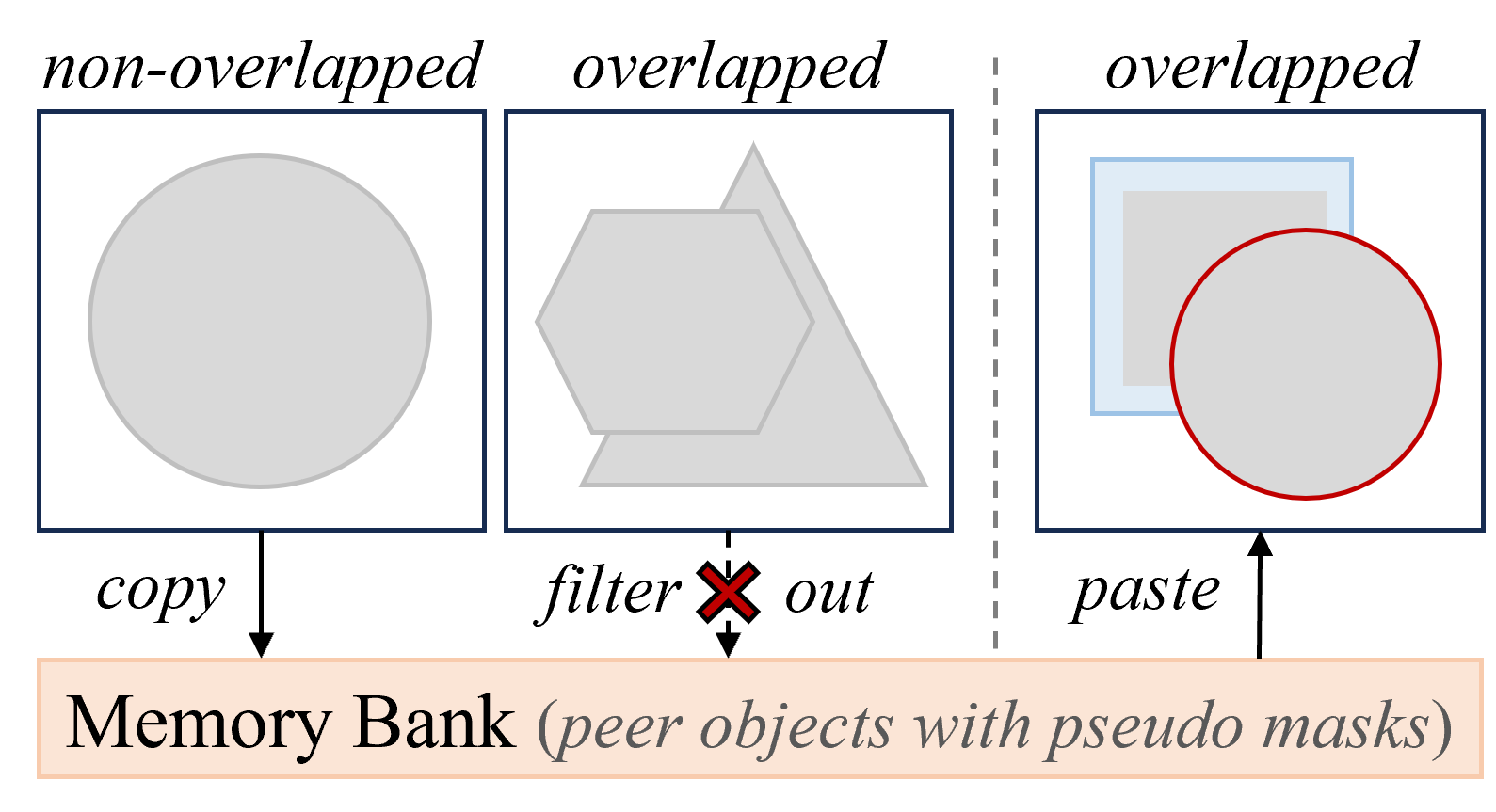}
\vspace{-5pt}
\caption{Peer-assisted Copy-paste: (1) Memory Bank collects the peer objects with high-quality pseudo masks non-overlapped with any objects. (2) The peer object is pasted into the image and overlapped with the object. The blue line is the low-quality mask of input object, and red line is the high-quality mask of peer object.}
\label{fig:pacp}
%\vspace{-5pt}
\end{figure}

\begin{algorithm}[tb]
\caption{Peer-assisted Copy-paste}
\label{alg:pc}
\begin{algorithmic}
\STATE \textbf{Input}: Input image $\mathcal{X}_i$, pseudo masks $m_i^t$, mask-quality scores $w_i$, instance segmentation model $f^{\theta}$, MemoryBank with length $H$, training iterations $E$, threshold $\tau$ for selecting peer tutors.

\STATE \textbf{Output}: Trained instance segmentation model $f^{\theta}$.

\STATE \textbf{Initialize Memory Bank}: $\text{MemoryBank} \leftarrow \emptyset$.

\STATE \textbf{Train Model}:
\FOR{$e = 1$ to $E$}
    \STATE \textbf{Update Memory Bank for Collecting Peer Tutors}:
    \FOR{each input image $\mathcal{X}_i$ with pseudo mask $m_i^t$ and score $w_i$}
        \IF{$w_i > \tau$ and $m_i^t$ is non-overlapped with any objects}
            \STATE Add Peer Tutor $\{ \mathcal{X}_i, m_i^t, w_i \}$ to $\text{MemoryBank}$.
            \IF{$|\text{MemoryBank}| > H$}
                \STATE Pop the oldest entry from $\text{MemoryBank}$.
            \ENDIF
        \ENDIF
    \ENDFOR

    \STATE
    \STATE \textbf{Copy-paste Operation}:
    \FOR{each input image $\mathcal{X}_i$ and pseudo mask $m_i^t$}
        \STATE Randomly select a Peer Tutor $(\mathcal{X}_j, m_j^t, w_j)$ in Memory Bank.
        \STATE Copy $\mathcal{X}_j \odot m_j^t$ and paste it into $\mathcal{X}_i$, ensuring overlap with the input object (i.e., the learner).
        \STATE Generate augmented image $\mathcal{X}_i^{\text{aug}}$ and corresponding mask $m_i^{\text{aug}}$.
    \ENDFOR

    \STATE
    \STATE \textbf{Model Training}: Compute loss on augmented data $(\mathcal{X}_i^{\text{aug}}, m_i^{\text{aug}})$, and update model $f^{\theta}$ using gradient descent.
\ENDFOR

\STATE \textbf{Return}: The trained instance segmentation model $f^{\theta}$.
\end{algorithmic}
\end{algorithm}

\subsection{Peer-assisted Copy-paste}
After obtaining the high-quality pseudo masks through the Quality-Aware Module, we further introduce the Peer-assisted Copy-paste inspired by Peer-Assisted Learning to improve the quality of the low-quality masks.
To imitate Peer-Assisted Learning, the Peer-assisted Copy-paste implements two steps, including Selecting Peer Tutors and Teaching Learners.
As presented in Fig.\ref{fig:pacp} and \cref{alg:pc}, the Peer-assisted Copy-paste implements the following steps:
(1) Selecting Peer Tutors: Construct a Memory Bank to dynamically update the peer objects (i.e., the peer tutors), which are selected based on the mask-quality score $w_i$. Along side the usage of the mask-quality score estimated by the Quality-Aware Module, we use additional prior knowledge to select the peer objects, i.e., only non-overlapping objects are selected.
Because the non-overlapping object makes it easier to predict the accurate mask by the model.
(2) Teaching Learners: Copy and paste the peer object into the input image, and require the peer object to overlap with the input object (i.e., the learner). The peer object has the more precise pseudo mask, thus the overlapping edge between the peer object and the input object is precise, i.e., making the overlapping edge of the input object easier to segment by the model. Besides, overlapping the peer object and the input object can improve the model's ability to distinguish between instance objects.
%\textit{Formal Algorithm and more analysis} are presented in \cref{sec:ana_pc}.
\textit{More analysis} is presented in \cref{sec:ana_pc}.

\begin{table*}[t]
\centering
\renewcommand{\tabcolsep}{9.5pt}
\renewcommand\arraystretch{1.1}
\small
\begin{tabular}{l|l|c|ccc|ccc}
\hline
\textbf{Method} & \textbf{Backbone} & \textbf{Schedule} & \textbf{AP} & \textbf{AP$_{50}$} & \textbf{AP$_{75}$} & \textbf{AP$_{s}$} & \textbf{AP$_{m}$} & \textbf{AP$_{l}$}\\
\hline
\multicolumn{8}{l}{\textit{Mask-supervised methods}} \\
\hline
Mask R-CNN \cite{HeGDG17} & ResNet-50-FPN & $1\times$ & $35.5$ & $57.0$ & $37.8$ & $19.5$ & $37.6$ & $46.0$ \\
CondInst \cite{TianSC20} & ResNet-50-FPN & $1\times$ & $35.9$ & $57.0$ & $38.2$ & $19.0$ & $38.6$ & $46.7$ \\
CondInst \cite{TianSC20} & ResNet-50-FPN & $3\times$ & $37.7$ & $58.9$ & $40.3$ & $20.4$ & $40.2$ & $48.9$ \\
CondInst \cite{TianSC20} & ResNet-101-FPN & $3\times$ & $39.1$ & $60.9$ & $42.0$ & $21.5$ & $41.7$ & $50.9$ \\
%SOLO \cite{SOLOWangKSJL20} & ResNet-101-FPN & $6\times$ & $37.8$ & $59.5$ & $40.4$ & $16.4$ & $40.6$ & $54.2$ \\
%SOLOv2 \cite{SOLOWangKSJL20}  & ResNet-101-FPN & $6\times$ & $39.7$ & $60.7$ & $42.9$ & $17.3$ & $42.9$ & $57.4$ \\
\hline
\hline
\multicolumn{8}{l}{\textit{Box-supervised methods}} \\
\hline
BoxTeacher \cite{cheng2023boxteacher} & ResNet-50-FPN & $1\times$ & $32.9$ & $\textbf{54.1}$ & $34.2$ & $17.4$ & $36.3$ & $43.7$ \\
\textbf{BoxSeg (Ours)} & ResNet-50-FPN & $1\times$ & $\textbf{33.3}$ & $53.4$ & $\textbf{35.0}$ & $\textbf{17.4}$ & $\textbf{36.7}$ & $\textbf{45.3}$ \\
\hline
BoxInst \cite{BoxInst} & ResNet-50-FPN & $3\times$ & $32.1$ & $55.1$ & $32.4$ & $15.6$ & $34.3$ & $43.5$\\
%DiscoBox \cite{DiscoBox} & ResNet-50-FPN & $3\times$ & $32.0$ & $53.6$ & $32.6$ & $11.7$ & $33.7$ & $\textbf{48.4}$ \\
Box2Mask-C \cite{li2024box2mask} & ResNet-50-FPN & $3\times$ & 32.6 & 55.4 & 33.4 & 14.7 & 35.8 & 45.9  \\
BoxTeacher \cite{cheng2023boxteacher} & ResNet-50-FPN & $3\times$ & $35.0$ & $56.8$ & $36.7$ & $19.0$ & $38.5$ & $45.9$\\
\textbf{BoxSeg (Ours)} & ResNet-50-FPN & $3\times$ & $\textbf{35.7}$ & $\textbf{56.8}$ & $\textbf{37.7}$ & $\textbf{19.6}$ & $\textbf{39.6}$ & $\textbf{47.5}$ \\
\hdashline
Box2Mask-T \cite{li2024box2mask} & ResNet-50-FPN & $3\times$ & 36.7 & 61.9 & 37.2 & 18.2 & 39.6 & 53.2  \\
\textbf{BoxSeg-T (Ours)} & ResNet-50-FPN & $3\times$ & $\textbf{39.3}$ & $\textbf{63.2}$ & $\textbf{41.3}$ & $\textbf{21.6}$ & $\textbf{42.2}$ & $\textbf{55.3}$ \\
\hline
%BBTP \cite{BBTP} & ResNet-101-FPN & $1\times$ & $21.1$ & $45.5$ & $17.2$ & $11.2$ & $22.0$ & $29.8$ \\
%BBAM \cite{BBAM} & ResNet-101-FPN & $1\times$ & $25.7$ & $50.0$ & $23.3$ & - & - & - \\
%BoxCaseg \cite{Boxcaseg} & ResNet-101-FPN & $1\times$ & $30.9$ & $54.3$ & $30.8$ & $12.1$ & $32.8$ & $46.3$ \\
BoxInst \cite{BoxInst} & ResNet-101-FPN & $3\times$ & $33.2$ & $56.5$ & $33.6$ & $16.2$ & $35.3$ & $45.1$\\
%BoxLevelSet \cite{BoxLevelSet} & ResNet-101-FPN & $3\times$ & $33.4$ & $56.8$ & $34.1$ & $15.2$ & $36.8$ & $46.8$ \\
Box2Mask-C \cite{li2024box2mask} & ResNet-101-FPN & $3\times$ & 34.2 & 57.8 & 35.2 & 16.0 & 37.7 & 48.3 \\
SIM \cite{li2023sim} & ResNet-101-FPN & $3\times$ & $35.3$ & $58.9$ & $36.4$ & $18.4$ & $38.0$ & $47.5$ \\
BoxTeacher \cite{cheng2023boxteacher} & ResNet-101-FPN & $3\times$ & $36.5$ & $59.1$ & $38.4$ & $20.1$ & $40.2$ & $47.9$ \\
\textbf{BoxSeg (Ours)} & ResNet-101-FPN & $3\times$ & $\textbf{37.5}$ & $\textbf{59.4}$ & $\textbf{39.8}$ & $\textbf{20.8}$ & $\textbf{41.6}$ & $\textbf{49.9}$ \\
\hdashline
Box2Mask-T \cite{li2024box2mask} & ResNet-101-FPN & $3\times$ & 38.3 & 65.1 & 38.8 & 19.3 & 41.7 & 55.2 \\
\textbf{BoxSeg-T (Ours)} & ResNet-101-FPN & $3\times$ & $\textbf{40.9}$ & $\textbf{65.6}$ & $\textbf{43.2}$ & $\textbf{22.6}$ & $\textbf{44.1}$ & $\textbf{57.5}$ \\
\hline
%BoxLevelSet \cite{BoxLevelSet} & ResNet-101-DCN-FPN & $3\times$ & $35.4$ & $59.1$ & $36.7$ & $16.8$ & $38.5$ & $51.3$ \\
DiscoBox \cite{DiscoBox} & ResNet-101-DCN-FPN & $3\times$ & $35.8$ & $59.8$ & $36.4$ & $16.9$ & $38.7$ & $52.1$ \\
SIM \cite{li2023sim} & ResNet-101-DCN-FPN & $3\times$ & $37.4$ & $\textbf{61.8}$ & $38.6$ & $18.6$ & $40.2$ & $51.6$ \\
BoxTeacher \cite{cheng2023boxteacher} & ResNet-101-DCN-FPN & $3\times$ & $37.6$ & $60.3$ & $39.7$ & $21.0$ & $41.8$ & $49.3$\\
\textbf{BoxSeg (Ours)} & ResNet-101-DCN-FPN & $3\times$ & $\textbf{38.6}$ & $60.9$ & $\textbf{40.9}$ & $\textbf{21.4}$ & $\textbf{42.2}$ & $\textbf{52.2}$ \\
% \hdashline
% SIM \cite{li2023sim} & Swin-B &  &  &  &  &  &  &  \\
% Box2Mask \cite{li2024box2mask} & Swin-B &  &  &  &  &  &  &  \\
% \textbf{BoxSeg (Ours)} & Swin-B &  &  &  &  &  &  &  \\
\hline
\end{tabular}
%\vspace{-5pt}
\caption{Comparisons between BoxSeg and state-of-the-arts on COCO \texttt{test-dev} for instance segmentation.}
\label{tab:main_experiments}
\vspace{-5pt}
\end{table*}

\begin{table}[t]
\centering
\renewcommand{\tabcolsep}{5.7pt}
\renewcommand\arraystretch{1.1}
\small
\begin{tabular}{l|l|ccc}
\hline
\textbf{Method} & \textbf{Backbone} & \textbf{AP} & \textbf{AP$_{50}$} & \textbf{AP$_{75}$} \\
\hline
BoxInst \cite{BoxInst} & ResNet-50 & $34.3$ & $59.1$ & $34.2$ \\
%DiscoBox \cite{DiscoBox} & ResNet-50 & - & $59.8$ & $35.5$ \\
%BoxLevelSet \cite{BoxLevelSet} & ResNet-50 & $36.3$ & $64.2$ & $35.9$\\
SIM \cite{li2023sim} & ResNet-50 & 36.7 & 65.5 & 35.6 \\
Box2Mask-C \cite{li2024box2mask} & ResNet-50 & 38.0 & 65.9 & 38.2 \\
BoxTeacher \cite{cheng2023boxteacher} & ResNet-50 & $38.6$ & $66.4$ & $38.7$ \\
\textbf{BoxSeg (Ours)} & ResNet-50 & \textbf{39.5} & \textbf{66.9} & \textbf{39.8} \\
\hdashline
Box2Mask-T \cite{li2024box2mask} & ResNet-50 & 41.4 & 68.9 & 42.1 \\
\textbf{BoxSeg-T (Ours)} & ResNet-50 & \textbf{42.4} & \textbf{69.4} & \textbf{43.2} \\
\hline
%BBTP \cite{BBTP} & ResNet-101 & 27.5 & $58.9$ & $21.6$ \\ 
%BBAM \cite{BBAM} & ResNet-101 & - & $63.7$ & $31.8$ \\
BoxInst \cite{BoxInst} & ResNet-101 & $36.4$ & $61.4$ & $37.0$ \\
%DiscoBox \cite{DiscoBox} & ResNet-101 & - & $62.2$ & $37.5$ \\
%BoxLevelSet \cite{BoxLevelSet} & ResNet-101 & $38.3$ & $66.3$ & $38.7$\\
SIM \cite{li2023sim} & ResNet-101 & $38.6$ & $67.1$ & $38.3$ \\
Box2Mask-C \cite{li2024box2mask} & ResNet-101 & 39.6 & 66.6 & 40.9 \\
BoxTeacher \cite{cheng2023boxteacher} & ResNet-101 & $40.3$ & $67.8$ & $41.3$ \\
\textbf{BoxSeg (Ours)} & ResNet-101 & \textbf{41.5} & \textbf{68.6} & \textbf{42.8} \\
\hdashline
Box2Mask-T \cite{li2024box2mask} & ResNet-101 & 43.2 & 70.8 & 44.4 \\
\textbf{BoxSeg-T (Ours)} & ResNet-101 & \textbf{44.8} & \textbf{71.6} & \textbf{45.2} \\
\hline
\end{tabular}
%\vspace{-5pt}
\caption{Comparisons on PASCAL VOC \texttt{val} for box-supervised instance segmentation.}
\label{tab:main_voc_experiments}
\vspace{-5pt}
\end{table}

\begin{table}[t]
\centering
\renewcommand{\tabcolsep}{4.0pt}
\renewcommand\arraystretch{1.1}
\small
\begin{tabular}{cccc|cc|cc}
\hline
\multicolumn{4}{c|}{\textbf{BoxTeacher}} & \multicolumn{2}{c|}{\textbf{BoxSeg}} & \multicolumn{2}{c}{\textbf{AP}} \\
\hline
$\mathcal{L}_{\text{pixel}}$ &BPMA & $\mathcal{L}_{\text{affinity}}$ & MCS & QAM& PC & AP$_{1\times}$ & AP$_{3\times}$ \\
\hline
\gt{\checkmark} & \gt{-} & \gt{-} & \gt{-} & \gt{-} & \gt{-} &\gt{31.0} &\gt{32.5}\\
\checkmark & \checkmark  & \checkmark & \checkmark & - &- &32.6 &34.2\\
\hdashline
\checkmark  & - &- & - & \checkmark &- &33.0 &34.8\\
\checkmark  &- & - & - & \checkmark &\checkmark &\textbf{33.1} &\textbf{35.6}\\
\hline
\end{tabular}
%\vspace{-5pt}
\caption{Comparisons on COCO \texttt{val} between our BoxSeg and BoxTeacher with ResNet-50. The main components are as follows: BPMA (Box-based Pseudo Mask Assignment), $\mathcal{L}_{\text{pixel}}$ (Pixel-wise segmentation loss), $\mathcal{L}_{\text{affinity}}$ (noise-reduced mask Affinity loss), MCS (Mask-aware Confidence Score), and our QAM (Quality-Aware Module), PC (Peer-assisted Copy-paste). AP$_{1\times}$ and AP$_{3\times}$ denote the AP with learning schedules of $1\times$ and $3\times$, respectively.}
\vspace{-5pt}
\label{tab:ablation_boxseg}
\end{table}

\begin{table}
\centering
\renewcommand{\tabcolsep}{7.5pt}
\renewcommand\arraystretch{1.1}
\small
\begin{tabular}{c|ccc|cc}
\hline
\multicolumn{1}{c|}{\textbf{BoxTeacher}} & \multicolumn{3}{c|}{\textbf{BoxSeg}} & \multicolumn{2}{c}{\textbf{AP}} \\
\hline
BPMA & BQR & QMF & MQS & AP$_{1\times}$ & AP$_{3\times}$ \\
\hline
\gt{-} & \gt{-} & \gt{-} & \gt{-} & \gt{31.0} & \gt{32.5} \\
\checkmark & - & - & - & 31.8 & 33.5 \\
- & \checkmark  & - & - & 32.1 & 33.8\\
- & \checkmark  & \checkmark & - & 32.7 & 34.5\\
- & \checkmark  & \checkmark & \checkmark & \textbf{33.0} & \textbf{34.8} \\
\hline
\end{tabular}
%\vspace{-5pt}
\caption{Effect of Quality-Aware Module on COCO \texttt{val} under BoxSeg framework with ResNet-50 without Peer-assisted Copy-paste. The Quality-Aware Module consists of BQR (Box-Quality Ranking), QMF (Quality-aware Masks Fusion), and MQS (Mask-Quality Scoring).}
\vspace{-5pt}
\label{tab:ablation_qam}
\end{table}

\section{Experiments}
\textbf{Datasets.} Following \cite{BoxInst,cheng2023boxteacher,li2023sim}, the COCO \cite{COCOLinMBHPRDZ14} and PASCAL VOC \cite{pascalvoc} datasets, are used for evaluation.
TThe COCO dataset comprises $80$ classes and includes $110k$ images for training, $5k$ for validation, and $20k$ for testing. The PASCAL VOC contains $20$ classes with $10582$ and $1449$ images for training and validation. 
In box-supervised instance segmentation task, we utilize only the bounding boxes and disregard the segmentation masks during training.

\noindent\textbf{Data Augmentation.}
The images fed into the teacher model are fixed to $800\times1333$ without any perturbation.
For the images input to the student model, we employ multi-scale augmentation and random horizontal flipping, which randomly resizes images between $640$ to $800$. Additionally, we randomly adopt color jittering, grayscale, and Gaussian blur for stronger augmentation. 

\noindent\textbf{Implementation Details.}
The proposed BoxSeg is implemented with {\texttt{Detectron2}}~\cite{wu2019detectron2} and trained with $8$ GPUs.
CondInst \cite{TianSC20} is adopted as the basic instance segmentation method, and the backbone with FPN is pre-trained on ImageNet.
Unless specified, we adopt the $3\times$ learning schedule ($270k$ iterations) with the SGD optimizer and the initial learning rate $0.01$.
The momentum used to update the teacher model is set to 0.999.
Empirically, the number of candidate boxes $K$ for Quality-Aware Module is set to 10, and the length of memory bank for Peer-assisted Copy-paste is set to 80.

\subsection{Comparisons with State-of-the-Arts}
We compare the proposed BoxSeg with the state-of-the-art box-supervised methods on COCO \texttt{test-dev} and PASCAL VOC \texttt{val}.
As illustrated in Tab.\ref{tab:main_experiments} and Tab.\ref{tab:main_voc_experiments}, our BoxSeg achieves new state-of-the-art results under different backbones and training schedules.
On the challenging COCO \texttt{test-dev} dataset, under ResNet-101-FPN backbone, our BoxSeg outperforms BoxInst and BoxTeacher by $4.3$ and $1.0$ AP.
Impressively, our BoxSeg obtains remarkable results on large objects, significantly outperforming BoxInst and BoxTeacher by $4.8$ AP and $2.0$ AP with ResNet-101-FPN. 
This improvement is due to the integrated Quality-Aware Module and Peer-assisted Copy-past in teacher-student framework, which obtains high-quality pseudo masks for the model optimization.
In addition, the performance gap between our box-supervised method BoxSeg and the mask-supervised method CondInst, is reduced to $1.6$ AP with ResNet-101-FPN.

Besides, similar to Box2Mask-T \cite{li2024box2mask}, we also develop BoxSeg-T, which adopts a stronger transformer-based instance segmentation decoder inspired by MaskFormer \cite{maskformer}. The results show that our BoxSeg-T achieves higher performance, indicating that using a stronger instance segmentation model as the basic model leads to a better performance on box-supervised instance segmentation.

\subsection{Ablation Study}
\textbf{BoxSeg \textit{vs.} BoxTeacher.}
In the 1st row of Tab.\ref{tab:ablation_boxseg}, the baseline, only applying the box-supervised loss \cite{BoxInst} and $\mathcal{L}_{\text{pixel}}$, achieves $31.0$ AP$_{1\times}$ and $32.5$ AP$_{3\times}$.
In the 3rd row, our QAM achieves performance improvements of 2.0 AP$_{1\times}$ and 2.3 AP$_{3\times}$ compared to the baseline.
Besides, comparing the 2nd and 3rd rows, our QAM improves performance by 0.4 AP$_{1\times}$ and 0.6 AP$_{3\times}$ over the BoxTeacher, demonstrating that the proposed QAM obtains higher quality pseudo masks than BoxTeacher, resulting in better performance.
Furthermore, comparing the 3rd and 4th rows, applying our PC method achieves further improvements of 0.1 AP$_{1\times}$ and 0.8 AP$_{3\times}$, indicating that the proposed PC is effective in refining the masks, contributing to higher accuracy.
Overall, the combination of QAM and PC in our BoxSeg achieves the highest performance of 33.1 AP$_{1\times}$ and 35.6 AP$_{3\times}$, significantly outperforming the BoxTeacher.

\noindent\textbf{Effects of Quality-Aware Module.}
In the 2nd row of Tab.\ref{tab:ablation_qam}, BPMA achieves 31.8 AP$_{1\times}$ and 33.5 AP$_{3\times}$, showing a significant improvement over the baseline of the 1st row. Comparing the 2nd and 3rd rows, our BQR method, ensuring that high-quality pseudo masks are retained, is superior to BPMA. As shown in the 3rd and 4th rows, our QMF achieves significant improvements by 0.6 AP$_{1\times}$ and 0.7 AP$_{3\times}$, indicating that the proposed QMF is effective in obtaining more accurate pseudo masks. In the last row, the proposed MQS obtains further improvements, by accurately measuring the quality of predicted masks to reduce the impact of noisy masks.

\noindent\textbf{Effects of Peer-assisted Copy-paste.} 
In the 2nd row of Tab.\ref{tab:ablation_pc}, randomly selecting and pasting the peer objects achieves 35.0 AP$_{3\times}$, showing an improvement by 0.2 AP$_{3\times}$ over the baseline of the 1st row. Comparing the 2nd and 3rd rows, selecting the peer objects with high-quality pseudo masks by our PC method (i.e., Selecting Peer Tutors), achieves significant improvements by 0.4 AP$_{3\times}$.
In the last row, pasting the peer object into the input image with overlapped (i.e., Teaching Learners), achieves further improvement by 0.2 AP$_{3\times}$.
Overall, our PC method achieves significant improvements by 0.8 AP$_{3\times}$ over the baseline.

\noindent\textbf{Applicabilities of QAM and PC.}
As illustrated in Tab.\ref{tab:other}, integrating our PC and QAM with other box-supervised instance segmentation methods, achieves competitive performance improvements.
The QAM can be integrated into the SIM model (teacher-student framework), and PC can be applied to SIM and BoxInst (single-model framework).

\begin{table}
\centering
\renewcommand{\tabcolsep}{8pt}
\renewcommand\arraystretch{1.1}
\small
\begin{tabular}{cc|c}
\hline
\textbf{Select peer objects} & \textbf{Paste peer objects} & \textbf{AP$_{3\times}$} \\
\hline
\gt{-} & \gt{-} & \gt{34.8}  \\
random & random & 35.0  \\
high-quality & random  & 35.4 \\
high-quality & overlapped  & \textbf{35.6}  \\
\hline
\end{tabular}
%\vspace{-5pt}
\caption{Effect of Peer-assisted Copy-paste on COCO \texttt{val} under BoxSeg framework with ResNet-50.}
\vspace{-5pt}
\label{tab:ablation_pc}
\end{table}

\begin{table}[t]
\centering
\renewcommand{\tabcolsep}{9.5pt}
\renewcommand\arraystretch{1.1}
\small
\begin{tabular}{l|ccc}
\hline
\textbf{Method} & \textbf{AP} & \textbf{AP$_{50}$} & \textbf{AP$_{75}$} \\
\hline
BoxInst \cite{BoxInst}  & $33.2$ & $56.5$ & $33.6$  \\
BoxInst + PC  & $34.0$ & $56.9$ & $34.8$  \\
\hdashline
SIM \cite{li2023sim} & $35.3$ & $58.9$ & $36.4$ \\
SIM + PC  & $35.7$ & $59.2$ & $37.0$ \\
SIM + PC + QAM  & $36.1$ & $59.3$ & $37.6$ \\
\hdashline
\textbf{BoxSeg} & $\textbf{37.5}$ & $\textbf{59.4}$ & $\textbf{39.8}$ \\
\hline
\end{tabular}
%\vspace{-5pt}
\caption{Integrating our PC and QAM with other methods, evaluating on COCO \texttt{test-dev} under ResNet-101-FPN backbone.}
\label{tab:other}
\vspace{-5pt}
\end{table}

\begin{figure}[t]
\centering
\includegraphics[width=\linewidth]{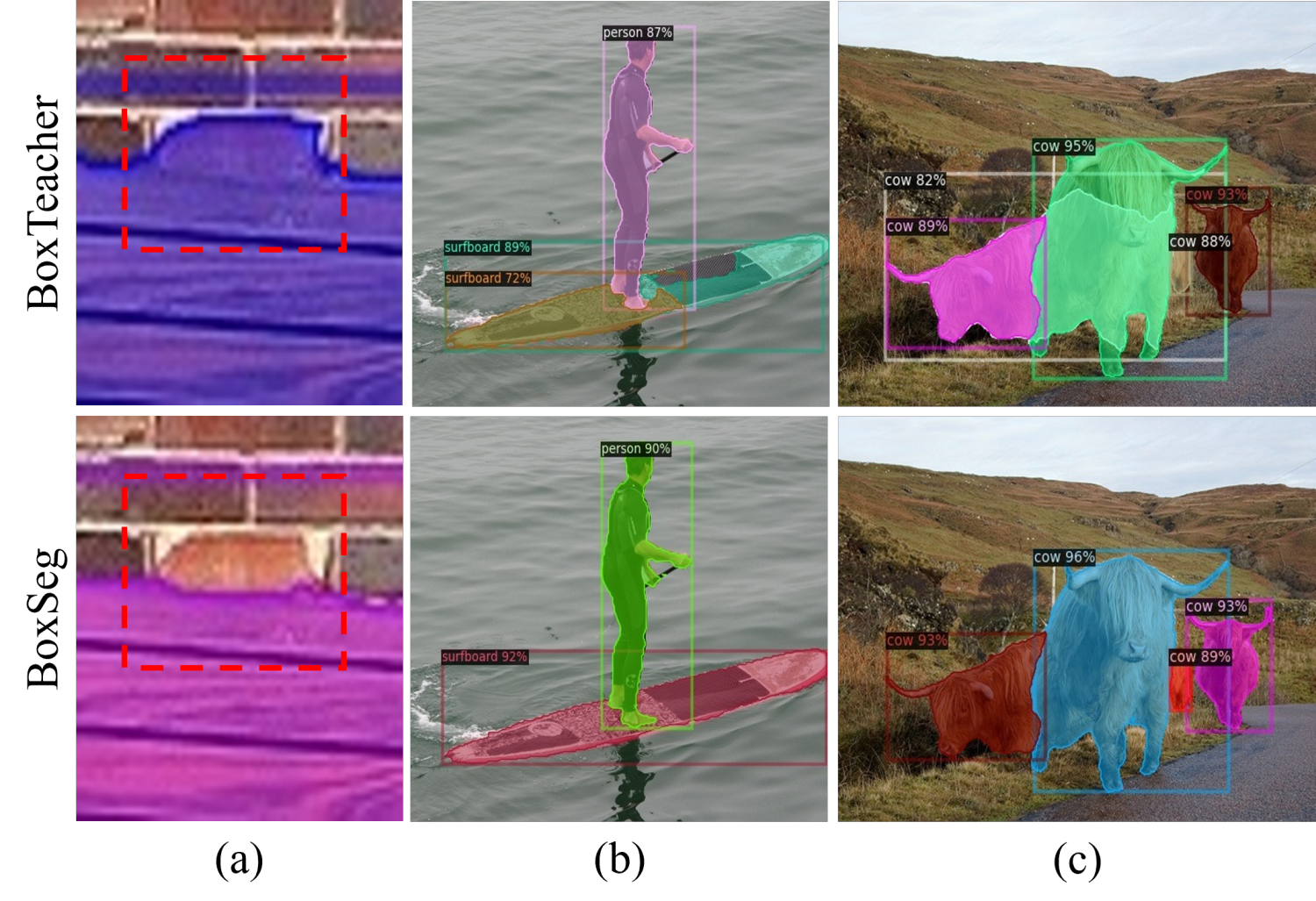}
\vspace{-25pt}
\caption{Visualization results of BoxTeacher (first row) and our BoxSeg (second row) with ResNet-101 on COCO \texttt{test-dev}. (a) Non-overlapped objects. (b) Overlapped inter-class objects. (c) Overlapped intra-class objects.}
\label{fig:vis}
\vspace{-12pt}
\end{figure}

\subsection{Qualitative Analyses}
Fig.\ref{fig:vis} shows the qualitative comparisons between our BoxSeg and BoxTeacher, and more visualizations are presented in Appendix.
Here we discuss three scenarios:

\noindent\textbf{Non-overlapped objects.} BoxTeacher struggles to distinguish similar backgrounds, while our BoxSeg excels at removing similar backgrounds. This can be attributed to the following reasons: BoxTeacher follows a one-to-one pattern, where one mask is predicted based on one box. However, due to the instability of predictions, it becomes challenging to distinguish similar backgrounds. In contrast, our Quality-aware Masks Fusion adopts a many-to-one mechanism, where one mask is predicted based on multiple boxes. Fusing predictions from boxes, enhances prediction stability, resulting in more accurate masks and better removal of similar backgrounds.

\noindent\textbf{Overlapped inter-class objects.} BoxTeacher produces redundant results, whereas our BoxSeg obtains more accurate masks. This can be explained by the following analysis: When a large object is occluded and divided into multiple parts, it is prone to be recognized as multiple distinct instances by the BoxTeacher, as it selects results based on a box-IOU threshold of 0.5. This leads to different parts of the same object being identified as separate instances. In contrast, our Box-Quality Ranking selects the top-${K}$ results based on box-IOU, enabling it to obtain more complete instances of large objects. Additionally, our Peer-assisted Copy-paste enhances the handling of occlusion scenarios by constructing data with overlapped instances. The results in Tab.\ref{tab:main_experiments} also demonstrate the significant performance advantage of our method on large objects, with an improvement of approximately $2$ AP across different backbone settings.

\noindent\textbf{Overlapped intra-class objects.} BoxTeacher produces redundant results, while our method achieves higher accuracy. This can be attributed to the following reasons: When intra-class objects are overlapped, the BoxTeacher may incorrectly merge different objects into a single instance. Leveraging the proposed Quality-Aware Module and Peer-assisted Copy-paste methods, our BoxSeg effectively distinguishes overlapped similar instances.

\section{Conclusions}
In this paper, we propose a BoxSeg framework for BSIS task, involving two novel modules named Quality-Aware Module (QAM) and Peer-assisted Copy-paste (PC).
The QAM leverages a quality-aware multi-mask complementation mechanism to generate high-quality pseudo masks, and effectively reduce the influence of noisy masks. 
Specifically, the QAM implements Box-Quality Ranking, Quality-aware Masks Fusion, and Mask-Quality Scoring, which ensures that high-quality pseudo masks are retained, obtain more accurate pseudo masks, and better measure the mask quality, respectively.
The PC module, inspired by peer-assisted learning, enhances the quality of low-quality masks by utilizing high-quality pseudo masks as guidance. Theoretical analyses and extensive experimental results demonstrate the effectiveness of the proposed modules, showing that BoxSeg outperforms state-of-the-art methods in BSIS task. Furthermore, the QAM and PC modules exhibit strong generalizability and can be seamlessly integrated into other frameworks to improve their performance.

\bibliographystyle{ACM-Reference-Format}
\bibliography{sample-base}

% APPENDIX
%%%%%%%%%%%%%%%%%%%%%%%%%%%%%%%%%%%%%%%%%%%%%%%%%%%%%%%%%%%%%%%%%%%%%%%%%%%%%%%
%%%%%%%%%%%%%%%%%%%%%%%%%%%%%%%%%%%%%%%%%%%%%%%%%%%%%%%%%%%%%%%%%%%%%%%%%%%%%%%
%\iffalse
%\newpage
\clearpage
\appendix

\section{More Experiments}
\textbf{Effect of Adaptive-$K$.}
To assess the effectiveness of the Adaptive-$K$ in QMF, we perform an ablation study comparing BoxSeg with and without it. The detailed results are presented in Tab.\ref{tab:adaptive_k}.
With the Adaptive-$K$ enabled, BoxSeg achieves a 0.3 improvement in overall AP, indicating more effective pseudo mask fusion. Notably, AP$_s$ improves by 0.6, showing better performance on small objects. Improvements in AP$_m$ and AP$_l$ also support the hypothesis that adaptively choosing the number of masks based on object scale enhances segmentation quality.

\begin{table}[ht]
\centering
\renewcommand{\tabcolsep}{4.5pt}
\renewcommand\arraystretch{1.1}
\small
\begin{tabular}{l|ccc|ccc}
\hline
\textbf{Method} & \textbf{AP} & \textbf{AP$_{50}$} & \textbf{AP$_{75}$} & \textbf{AP$_{s}$} & \textbf{AP$_{m}$} & \textbf{AP$_{l}$}\\
\hline
\textbf{w/o Adaptive-$K$} & 37.2 & 59.2 & 39.4 & 20.2 & 41.4 & 49.7 \\
\textbf{w/ Adaptive-$K$} & $\textbf{37.5}$ & $\textbf{59.4}$ & $\textbf{39.8}$ & $\textbf{20.8}$ & $\textbf{41.6}$ & $\textbf{49.9}$ \\
\hline
\end{tabular}
%\vspace{-5pt}
\caption{Effect of Adaptive-$K$ in QMF, evaluating on COCO \texttt{test-dev} under ResNet-101-FPN backbone.}
\label{tab:adaptive_k}
\vspace{-5pt}
\end{table}

\noindent\textbf{\textbf{BoxSeg \textit{vs.} BoxTeacher+CopyPaste.}}
Tab.\ref{tab:copy_experiments} demonstrates that BoxSeg consistently outperforms BoxTeacher and its enhanced variant with standard CopyPaste augmentation across both ResNet-50-FPN and ResNet-101-FPN backbones. 
While the CopyPaste strategy brings marginal gains to BoxTeacher (e.g., +0.2 AP with ResNet-101-FPN), the improvements from BoxSeg are notably larger. This clearly highlights the effectiveness of our Quality-Aware Module (QAM) and Peer-assisted Copy-paste (PC) module.

\begin{table}[ht]
\centering
\renewcommand{\tabcolsep}{4.5pt}
\renewcommand\arraystretch{1.1}
\small
\begin{tabular}{l|l|ccc}
\hline
\textbf{Method} & \textbf{Backbone} & \textbf{AP} & \textbf{AP$_{50}$} & \textbf{AP$_{75}$}\\
\hline
BoxTeacher \cite{cheng2023boxteacher} & ResNet-50-FPN& $35.0$ & $56.8$ & $36.7$\\
BoxTeacher+CopyPaste & ResNet-50-FPN& $35.1$ & 56.8 & 36.9 \\
\textbf{BoxSeg (Ours)} & ResNet-50-FPN & $\textbf{35.7}$ & $\textbf{56.8}$ & $\textbf{37.7}$ \\
\hline
BoxTeacher \cite{cheng2023boxteacher} & ResNet-101-FPN & $36.5$ & $59.1$ & $38.4$ \\
BoxTeacher+CopyPaste & ResNet-101-FPN & $36.7$ & 59.2 & 38.7 \\
\textbf{BoxSeg (Ours)} & ResNet-101-FPN & $\textbf{37.5}$ & $\textbf{59.4}$ & $\textbf{39.8}$ \\
\hline
\end{tabular}
%\vspace{-5pt}
\caption{Comparisons between BoxSeg and BoxTeacher on COCO \texttt{test-dev} for instance segmentation.}
\label{tab:copy_experiments}
\vspace{-5pt}
\end{table}

\section{Proof of Generalization Error Bound for Quality-aware Masks Fusion}
\label{sec:ana_qmf}
\begin{assumption}
The candidate masks $m_{i,n}$ have errors $\epsilon_{m,n}$ such that $m_{i,n} = m_i^* + \epsilon_{m,n}$, where $m_i^*$ is the true mask and $\epsilon_{m,n}$ has zero mean and variance $\sigma_m^2$.
\end{assumption}
\begin{assumption}
The box-quality score $\sqrt{s_{i,n} \cdot u_{i,n}}$ have errors $\epsilon_{w,n}$ with zero mean and variance $\sigma_w^2$.
\end{assumption}
\begin{assumption}
The threshold $\tau_m$ filters out low-quality masks, and the indicator function $\mathbbm{1}(s_{i,n} > \tau_m)$ is applied.
\end{assumption}

\begin{proof} 

\textbf{Upper Bound}.
The upper bound is derived by decomposing the generalization error into three components:

\textit{Approximation Error}: This is bounded by the maximum deviation of the candidate masks from the true mask:
\begin{equation}
\text{Approximation Error} \leq \max_{n} \|m_{i,n} -  m_i^*\|.
\end{equation}

\textit{Estimation Error}: This arises from the finite number of effective candidate masks $\hat{K} = \sum_{n=1}^{K} \mathbbm{1}(s_{i,n} > \tau_m)$. Using Hoeffding’s inequality, we bound this as:
\begin{equation}
\text{Estimation Error} \leq O\left(\sqrt{\frac{\log(1/\delta)}{\hat{K}}}\right),
\end{equation}
with probability at least $1 \delta$.
   
\textit{Weighting Error}: This is due to errors in the box-quality score. Let $\epsilon_w$ be the maximum error in the weights. Then:
\begin{equation}
\text{Weighting Error} \leq \epsilon_w \cdot \max_{n} \|m_{i,n}\|.
\end{equation}

Combining these components, we obtain the upper bound:
\begin{equation}
\text{Error}(m_i^t) \leq \max_{n} \|m_{i,n} m_i^*\| + O\left(\sqrt{\frac{\log(1/\delta)}{\hat{K}}}\right) + \epsilon_w \cdot \max_{n} \|m_{i,n}\|.
\end{equation}

%\textbf{Lower Bound}.
%The lower bound is derived by considering the best-case scenario where the candidate masks and box-quality score are as accurate as possible. However, due to the finite number of candidate masks $K$, the error cannot be reduced below:
%\begin{equation}
%\text{Error}(m_i^t) \geq \Omega\left(\sqrt{\frac{1}{K}}\right).
%\end{equation}

This follows from the fact that the estimation error scales as $\sqrt{1/\hat{K}}$, even in the absence of other sources of error.

\end{proof}

\begin{proof} 

\textbf{Maximum Error $\epsilon_w$}.
To compute the maximum error $\epsilon_w$, we analyze how errors in the box-quality score propagate to the weights used in the QMF method.

\textit{Definition of Weights}.
The weights $w_n$ are computed as:
\begin{equation}
w_n = \frac{\mathbbm{1}(s_{i,n} > \tau_m) \cdot \sqrt{s_{i,n} \cdot u_{i,n}}}{\sum_{k=1}^K \mathbbm{1}(s_{i,k} > \tau_m) \cdot \sqrt{s_{i,k} \cdot u_{i,k}}},
\end{equation}
where,
$s_{i,n}$ is the box-score of the $n$-th candidate mask,
$u_{i,n}$ is the box-IOU between the predicted box and the ground truth (GT) box,
$\mathbbm{1}(s_{i,n} > \tau_m)$ is the indicator function that filters out masks with box-score below the threshold $\tau_m$.

\textit{Error in box-quality score}.
The box-quality score $\sqrt{s_{i,n} \cdot u_{i,n}}$ is subject to errors due to noise in box-score $s_{i,n}$ and noise in box-IOU $u_{i,n}$.
Let $\epsilon_{s,n}$ and $\epsilon_{u,n}$ be the errors in $s_{i,n}$ and $u_{i,n}$, respectively. The noisy box-quality score can be written as:
\begin{equation}
\sqrt{(s_{i,n} + \epsilon_{s,n}) \cdot (u_{i,n} + \epsilon_{u,n})}.
\end{equation}

\textit{First-Order Approximation of Error}.
Using a first-order Taylor approximation, the error in the box-quality score can be approximated as:
\begin{equation}
\sqrt{s_{i,n} \cdot u_{i,n}} + \epsilon_{w,n},
\end{equation}
where $\epsilon_{w,n}$ is the error in the box-quality score. The first-order approximation of $\epsilon_{w,n}$ is:
\begin{equation}
\epsilon_{w,n} \approx \frac{1}{2} \left( \frac{\epsilon_{s,n}}{s_{i,n}} + \frac{\epsilon_{u,n}}{u_{i,n}} \right) \cdot \sqrt{s_{i,n} \cdot u_{i,n}}.
\end{equation}

\textit{ Maximum Error in Weights}.
The maximum error in the weights $\epsilon_w$ is the largest possible deviation of the weights due to errors in the box-quality score. To compute $\epsilon_w$, we consider the worst-case scenario where the errors $\epsilon_{s,n}$ and $\epsilon_{u,n}$ are maximized.

\textit{Error Propagation in Weights}.
The weights $w_n$ are normalized, so the error in the weights depends on the errors in the numerator and denominator of the weight formula. Let ${w}_n^*$ be the true weight (without noise), $w_n$ be the noisy weight (with noise).
The error in the weights can be written as:
\begin{equation}
|w_n - {w}_n^*| \leq \epsilon_w.
\end{equation}

\textit{Bounding the Error}.
Using the first-order approximation of $\epsilon_{w,n}$, the maximum error in the weights $\epsilon_w$ can be bounded as:
\begin{equation}
\epsilon_w \leq \max_n \left| \frac{\partial w_n}{\partial \epsilon_{w,n}} \cdot \epsilon_{w,n} \right|.
\end{equation}
The partial derivative $\frac{\partial w_n}{\partial \epsilon_{w,n}}$ is:
\begin{equation}
\frac{\partial w_n}{\partial \epsilon_{w,n}} = \frac{1}{\sum_{k=1}^K \mathbbm{1}(s_{i,k} > \tau_m) \cdot \sqrt{s_{i,k} \cdot u_{i,k}}}.
\end{equation}
Thus, the maximum error in the weights $\epsilon_w$ is:
\begin{equation}
\epsilon_w \leq \max_n \left| \frac{\epsilon_{w,n}}{\sum_{k=1}^K \mathbbm{1}(s_{i,k} > \tau_m) \cdot \sqrt{s_{i,k} \cdot u_{i,k}}} \right|.
\end{equation}

\textit{Final Expression for $\epsilon_w$}.
Substituting the expression for $\epsilon_{w,n}$, we obtain:
\begin{equation}
\epsilon_w \leq \max_n \left| \frac{\frac{1}{2} \left( \frac{\epsilon_{s,n}}{s_{i,n}} + \frac{\epsilon_{u,n}}{u_{i,n}} \right) \cdot \sqrt{s_{i,n} \cdot u_{i,n}}}{\sum_{k=1}^K \mathbbm{1}(s_{i,k} > \tau_m) \cdot \sqrt{s_{i,k} \cdot u_{i,k}}} \right|.
\end{equation}
Simplifying, we get:
\begin{equation}
\epsilon_w \leq \frac{1}{2} \max_n \left| \frac{\epsilon_{s,n}}{s_{i,n}} + \frac{\epsilon_{u,n}}{u_{i,n}} \right| \cdot \frac{\sqrt{s_{i,n} \cdot u_{i,n}}}{\sum_{k=1}^K \mathbbm{1}(s_{i,k} > \tau_m) \cdot \sqrt{s_{i,k} \cdot u_{i,k}}}.
\end{equation}

\begin{remark}
The term $\frac{\epsilon_{s,n}}{s_{i,n}} + \frac{\epsilon_{u,n}}{u_{i,n}}$ represents the relative error in the box-quality score.
The term $\frac{\sqrt{s_{i,n} \cdot u_{i,n}}}{\sum_{k=1}^K \mathbbm{1}(s_{i,k} > \tau_m) \cdot \sqrt{s_{i,k} \cdot u_{i,k}}}$ represents the normalized contribution of the $n$-th mask to the weights.
The maximum error $\epsilon_w$ is proportional to the relative errors $\frac{\epsilon_{s,n}}{s_{i,n}}$ and $\frac{\epsilon_{u,n}}{u_{i,n}}$. Therefore, reducing these errors will minimize $\epsilon_w$.
Accurate box-quality score reduce $\epsilon_{s,n}$ and $\epsilon_{u,n}$, while thresholding $\tau_m$ filters out low-quality masks.
\end{remark}

\end{proof}

\section{Effectiveness Discussion of Quality-aware Masks Fusion}
The Quality-aware Masks Fusion (QMF) method is effective in generating high-quality pseudo masks due to its principled design and robust mechanisms. Below, we summarize the key factors contributing to its effectiveness:

$\bullet$ Combining Confidence and Spatial Alignment:
The QMF method leverages both the box-score $s_{i,n}$ (confidence in the prediction) and the box-IOU $u_{i,n}$ (spatial alignment with the ground truth box) to assess the quality of each candidate mask.
Masks with higher box-scores (greater confidence) and higher box-IOUs (better alignment) are assigned larger weights, ensuring that they have a greater influence on the final fused mask $m_i^t$.
This dual consideration of confidence and alignment ensures that the fused mask is both accurate and well-aligned with the true object.

$\bullet$ Balanced Fusion:
The weights $w_n$ are normalized such that they sum to 1, ensuring a balanced fusion process.
This normalization prevents any single mask from dominating the result and ensures that the final fused mask incorporates information from multiple candidate masks proportionally to their quality.

$\bullet$ Reducing the Impact of Low-Quality Masks:
The inclusion of a threshold $\tau_m$ in the indicator function $\mathbbm{1}(s_{i,n} > \tau_m)$ filters out low-confidence masks, preventing them from contributing to the fusion process.
Only high-confidence masks, which are more likely to be accurate, are included in the final fusion.
Additionally, masks with poor spatial alignment (low box-IOUs) receive smaller weights, minimizing their impact on the final result.

$\bullet$ Theoretical Guarantees:
The generalization error of the fused mask $m_i^t$ is bounded.
This bound ensures that the fused mask $m_i^t$ is accurate and robust to noise, with the error decreasing as the number of effective candidate masks $\hat{K}$ increases and the weighting error $\epsilon_w$ decreases.

\section{Proof of Estimation Error Bound for Mask-Quality Scoring}
\label{sec:ana_mqs}

\begin{proof}

\textit{Bounding the Estimation Error of $\hat{s}_i$}.
The box-score $\hat{s}_i$ is an empirical average of $\hat{K} = \sum_{n=1}^{K} \mathbbm{1}(s_{i,n} > \tau_m)$ i.i.d. random variables $s_{i,n}$ with mean $\mu_s$ and variance $\sigma_s^2$. By the Hoeffding inequality, we have:
\begin{equation}
\mathbb{P}\left(|\hat{s}_i - \mu_s| \geq \epsilon\right) \leq 2 \exp\left(-\frac{2\hat{K}\epsilon^2}{\sigma_s^2}\right).
\end{equation}

\textit{Bounding the Estimation Error of $\hat{m}_i^t$}.
The pixel-wise probability score $\hat{m}_i^t$ is an empirical average of $\hat{M} = \sum_{x,y}^{H,W} \mathbbm{1}(m_{i,(x,y)}^t > \tau_m)$ i.i.d. random variables $m_{i,(x,y)}^t$ with mean $\mu_m$ and variance $\sigma_m^2$. By the Hoeffding inequality, we have:
\begin{equation}
\mathbb{P}\left(|\hat{m}_i^t - \mu_m| \geq \epsilon\right) \leq 2 \exp\left(-\frac{2\hat{M}\epsilon^2}{\sigma_m^2}\right).
\end{equation}

\textit{Bounding the Estimation Error of $w_i$}.
The mask-quality score $w_i = \sqrt{\hat{s}_i \cdot \hat{m}_i^t}$ is the geometric mean of $\hat{s}_i$ and $\hat{m}_i^t$. Let $\mu_w = \sqrt{\mu_s \cdot \mu_m}$ be the expected value of $w_i$.
The estimation error $|w_i - \mu_w|$ depends on the errors of $\hat{s}_i$ and $\hat{m}_i^t$. Using the union bound, the probability that either $|\hat{s}_i - \mu_s| \geq \epsilon$ or $|\hat{m}_i^t - \mu_m| \geq \epsilon$ occurs is:
\begin{equation}
\mathbb{P}\left(|w_i - \mu_w| \geq \epsilon\right) \leq \mathbb{P}\left(|\hat{s}_i - \mu_s| \geq \epsilon\right) + \mathbb{P}\left(|\hat{m}_i^t - \mu_m| \geq \epsilon\right).
\end{equation}
Substituting the Hoeffding bounds for $\hat{s}_i$ and $\hat{m}_i^t$, we have:
\begin{equation}
\mathbb{P}\left(|w_i - \mu_w| \geq \epsilon\right) \leq 2 \exp\left(-\frac{2\hat{K}\epsilon^2}{\sigma_s^2}\right) + 2 \exp\left(-\frac{2\hat{M}\epsilon^2}{\sigma_m^2}\right).
\end{equation}

\textit{Connecting $\mu_w$ to True Mask Quality}.
The true mask quality is defined as $w_i^* = \text{IOU}(m_i^t, m_i^*)$.
Since $\mu_w = \sqrt{\mu_s \cdot \mu_m}$ approximates $w_i^*$, we can write:
\begin{equation}
|w_i - w_i^*| \leq |w_i - \mu_w| + |\mu_w - w_i^*|.
\end{equation}
For small $\epsilon$, the second term $|\mu_w - w_i^*|$ is negligible, and the bound on $|w_i - w_i^*|$ is dominated by $|w_i - \mu_w|$. This means that the estimation error of $w_i$ relative to $w_i^*$ is primarily determined by $|w_i - \mu_w|$, which we have already bounded using Hoeffding's inequality.

\textit{Final Bound on Estimation Error}.
Combining the bounds on $\hat{s}_i$ and $\hat{m}_i^t$, the estimation error $|w_i - w_i^*|$ is bounded with high probability:
\begin{equation}
\mathbb{P}\left(|w_i - w_i^*| \geq \epsilon\right) \leq 2 \exp\left(-\frac{2\hat{K}\epsilon^2}{\sigma_s^2}\right) + 2 \exp\left(-\frac{2\hat{M}\epsilon^2}{\sigma_m^2}\right).
\end{equation}

\textit{Convergence to True Mask Quality}.
As $\hat{K} \to \infty$ and $\hat{M} \to \infty$, the Hoeffding bounds imply that $\hat{s}_i \to \mu_s$ and $\hat{m}_i^t \to \mu_m$ almost surely. Therefore:
\begin{equation}
\lim_{\hat{K}, \hat{M} \to \infty} w_i = \sqrt{\mu_s \cdot \mu_m} = \mu_w \approx w_i^*.
\end{equation}

\end{proof}

\renewcommand\thefigure{6}
\begin{figure*}[h]
\centering
\includegraphics[width=\linewidth]{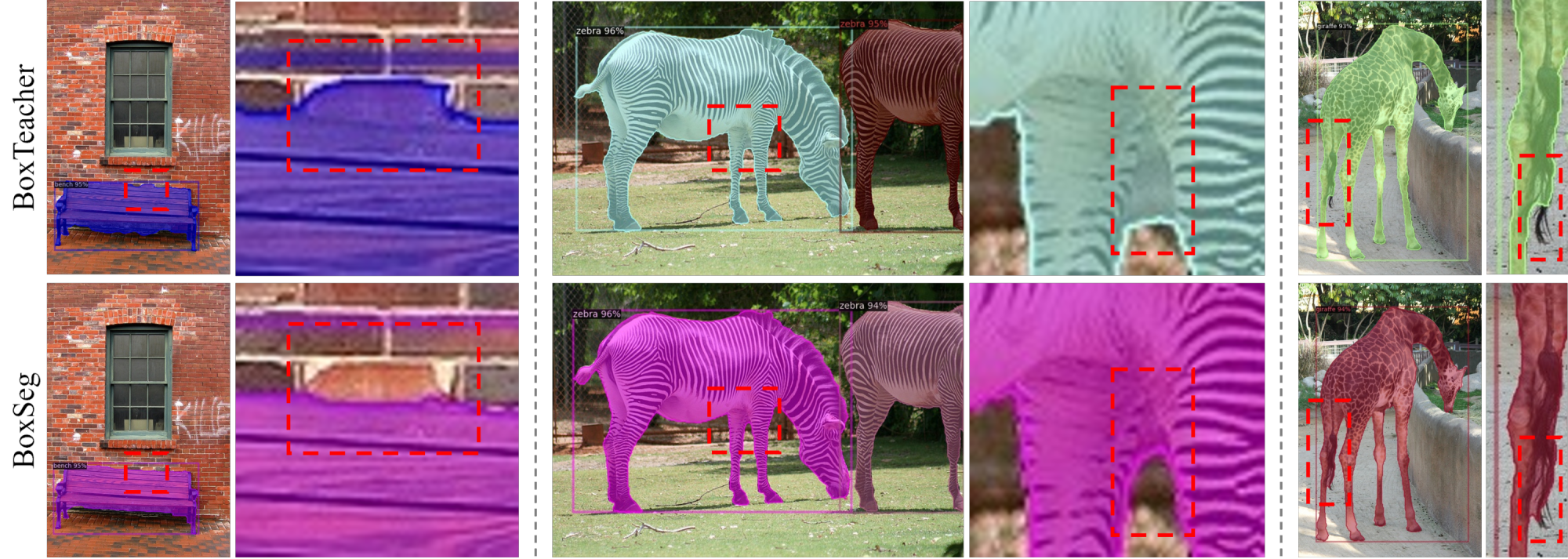}
%\vspace{-5pt}
\caption{\textbf{Non-overlapped objects}: visualizations of BoxTeacher and our BoxSeg with ResNet-101 on COCO \texttt{test-dev}.}
\label{fig:non_overlapped}
%\vspace{-5pt}
\end{figure*}

\renewcommand\thefigure{7}
\begin{figure*}[h]
\centering
\includegraphics[width=\linewidth]{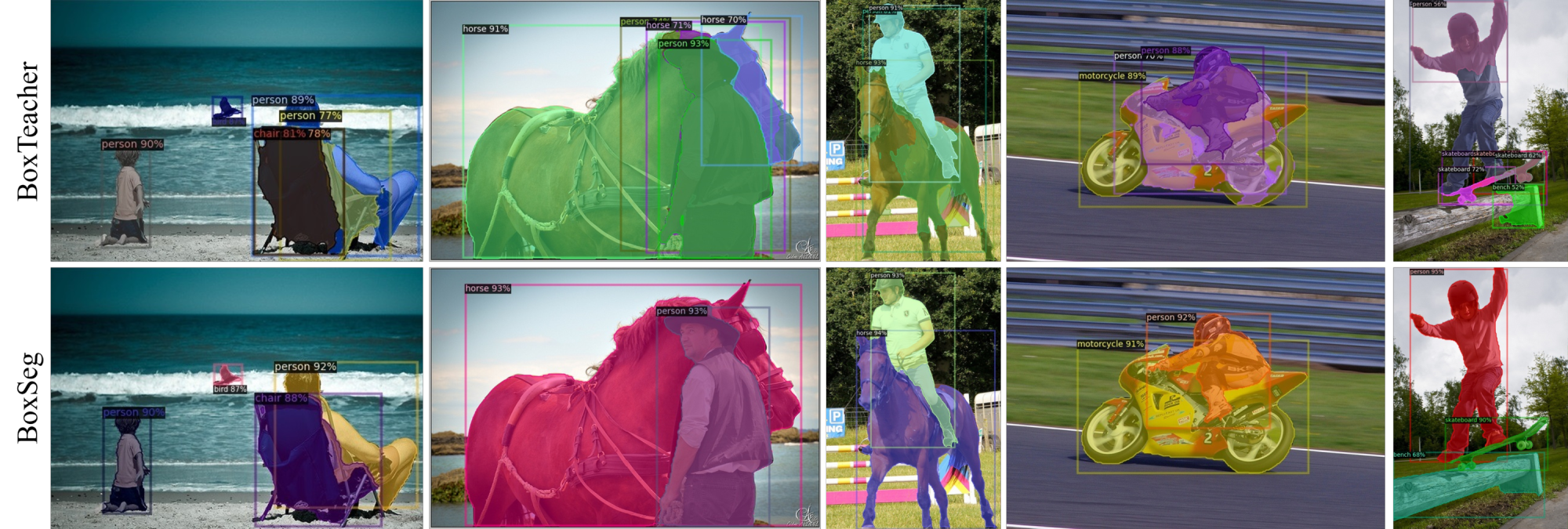}
%\vspace{-5pt}
\caption{\textbf{Overlapped inter-class objects}: visualizations of BoxTeacher and our BoxSeg with ResNet-101 on COCO \texttt{test-dev}.}
\label{fig:overlapped_interclass}
%\vspace{-5pt}
\end{figure*}

\renewcommand\thefigure{8}
\begin{figure*}[t]
\centering
\includegraphics[width=\linewidth]{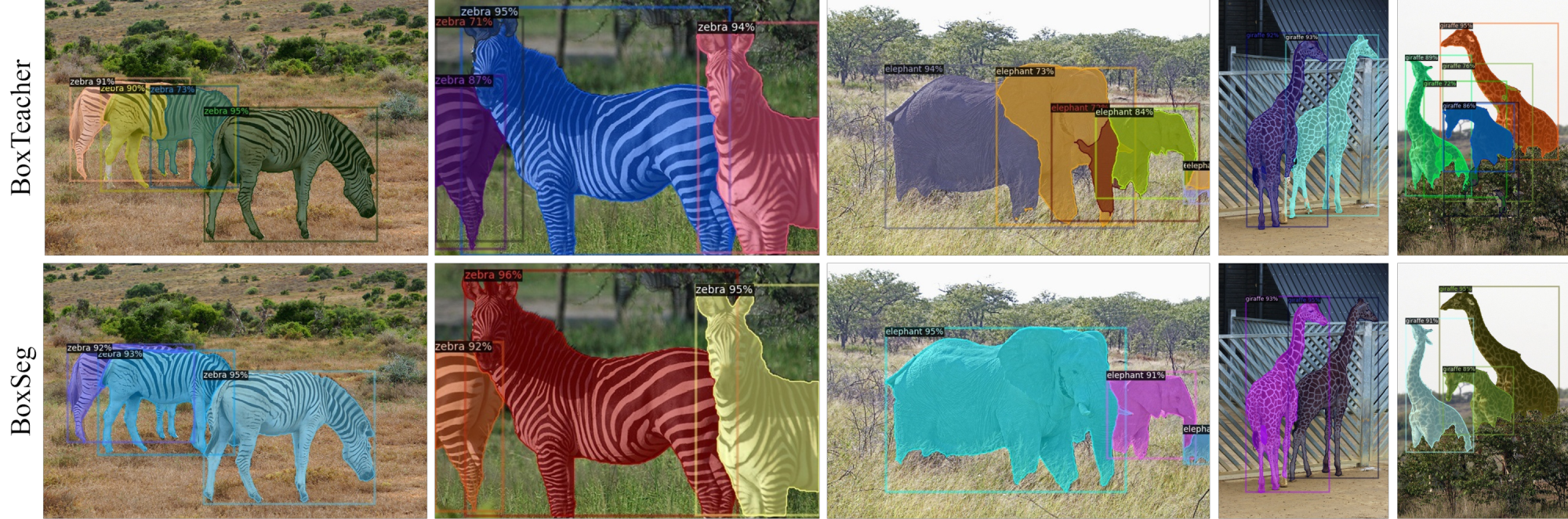}
%\vspace{-5pt}
\caption{\textbf{Overlapped intra-class objects}: visualizations of BoxTeacher and our BoxSeg with ResNet-101 on COCO \texttt{test-dev}.}
\label{fig:overlapped_intraclass}
%\vspace{-5pt}
\end{figure*}

\renewcommand\thefigure{9}
\begin{figure*}[t]
\centering
\includegraphics[width=\linewidth]{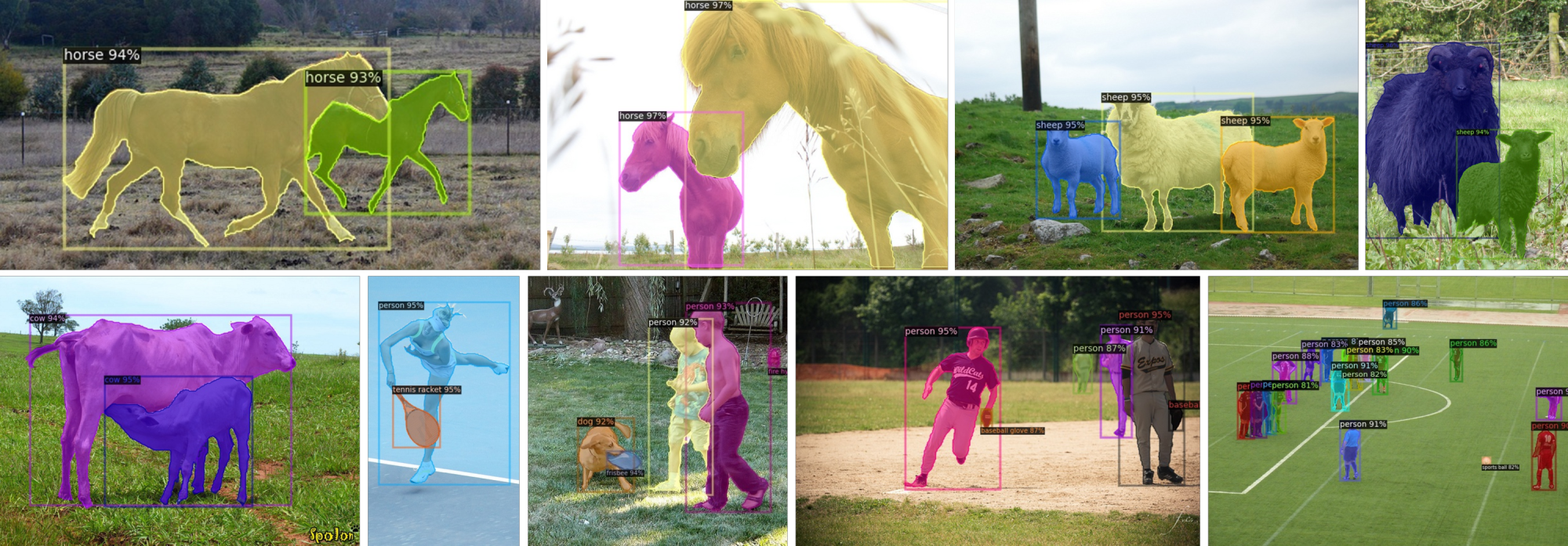}
%\vspace{-5pt}
\caption{\textbf{Overlapped objects}: visualizations of our BoxSeg with ResNet-101 on COCO \texttt{test-dev}.}
\label{fig:overlapped}
%\vspace{-5pt}
\end{figure*}

\section{Effectiveness Discussion of Mask-Quality Scoring}
We will show that the Mask-Quality Scoring effectively estimates the quality of the pseudo mask by considering both global and local information and how this combination leads to a robust and effective measure.

$\bullet$ Global Quality Component:
The box-score $s_i$ measures how well the predicted box aligns with the true object at a global level. In practical terms, this reflects the overall containment of the mask within the object and the confidence that the predicted object is correct. The higher the average box-score $\hat{s}_i$, the more likely it is that the pseudo mask $m_i^t$ is of high global quality.

$\bullet$ Local Quality Component:
The average pixel-wise probability score $\hat{m}_i^t$ ensures that the local quality of the mask is properly considered. This component focuses on the pixel-level accuracy and filters out noisy regions. The higher the average pixel-wise probability scores, the more confident the mask is in its alignment with the object at a local level.

$\bullet$ Combining Global and Local Information:
The combination of $\hat{s}_i$ and $\hat{m}_i^t$ in the form of $w_i = \sqrt{\hat{s}_i \cdot \hat{m}_i^t}$ provides a comprehensive quality measure:
(1) Higher $\hat{s}_i$: Indicates that the pseudo mask is globally aligned with the true object.
(2) Higher $\hat{m}_i^t$: Indicates that the mask has high local confidence in the pixels within the object region.
The product of these two components ensures that both aspects are given equal importance in determining the mask quality. If either component is low (e.g., low confidence in the box or low pixel probability), the mask-quality score $w_i$ will also be low, indicating a poor-quality mask.

$\bullet$ Noise Reduction and Robustness:
By including the thresholding mechanisms $\mathbbm{1}(s_{i,n} > \tau_m)$ and $\mathbbm{1}(m_{i,(x,y)}^t > \tau_m)$, Mask-Quality Scoring effectively filters out noisy masks and low-confidence pixels. This increases the robustness of the scoring mechanism:
(1) Global Thresholding: Ensures that only masks with high box-scores contribute to $\hat{s}_i$, reducing the influence of low-confidence masks.
(2) Local Thresholding: Ensures that only high-confidence pixels within the GT box are considered for $\hat{m}_i^t$, reducing the influence of noise outside the true object region.

$\bullet$ Theoretical Guarantees:
The effectiveness of Mask-Quality Scoring is supported by theoretical guarantees. The estimation error of $w_i$ is bounded with high probability. This ensures that $w_i$ is a reliable measure of mask quality, even for finite $\hat{K}$ and $\hat{M}$.

\section{Analysis for Quality-Aware Mask-Supervised Loss}
\label{sec:ana_qml}
We aim to show that the Quality-Aware Mask-Supervised Loss ($\mathcal{L}_{mask}$) is effective in focusing on high-quality pseudo masks while mitigating the influence of noisy masks. 
The core principle of $\mathcal{L}_{mask}$ is that the weight $w_i$ adapts the loss according to the quality of the pseudo mask, with higher-quality masks contributing more to the loss and lower-quality masks contributing less.
The score $w_i$ is designed to reflect both the global reliability of the pseudo mask (from the average box-score $\hat{s}_i$) and its local accuracy (from the average pixel-wise probability $\hat{m}_i^t$).

\paragraph{Emphasizing High-Quality Pseudo Masks.}
If the pseudo mask $m_i^t$ is of high quality, then both $\hat{s}_i$ and $\hat{m}_i^t$ will be large, leading to a larger weight $w_i$. This means that the pixel-wise loss for this mask will have a larger contribution to the total loss, which encourages the Student Model to focus on learning from high-quality pseudo labels.
Formally, if $\hat{s}_i$ and $\hat{m}_i^t$ are large, then:
\begin{equation}
w_i = \sqrt{\hat{s}_i \cdot \hat{m}_i^t} \quad \text{is large},
\end{equation}
which implies $\mathcal{L}_{mask}$ emphasizes this mask more.
This prioritization of high-quality pseudo masks helps improve the model’s learning by leveraging reliable pseudo labels, as these masks better align with the ground truth.

\paragraph{Reducing the Impact of Noisy Masks.}
If the pseudo mask $m_i^t$ is noisy or low-quality, then either $\hat{s}_i$ or $\hat{m}_i^t$, or both, will be small. As a result, $w_i$ will be small, reducing the influence of that particular mask on the total loss. This mechanism ensures that noisy or poorly predicted masks do not disrupt the training process.
Formally, if $\hat{s}_i$ or $\hat{m}_i^t$ is small:
\begin{equation}
w_i = \sqrt{\hat{s}_i \cdot \hat{m}_i^t} \quad \text{is small}, 
\end{equation}
leading to $\mathcal{L}_{mask}$ downweighting this mask.
This down-weighting prevents noisy pseudo masks from dominating the learning process, allowing the model to focus on more reliable signals.

\section{Analysis for Peer-assisted Copy-paste} 
\label{sec:ana_pc}
Peer-assisted Copy-paste (PC) leverages high-quality pseudo masks to improve the quality of low-quality masks.
It is a data augmentation technique that can improve model generalization by leveraging high-quality pseudo-labels. Through data augmentation, improved boundary accuracy, and enhanced instance discrimination, PC effectively reduces generalization errors. Here's a detailed breakdown of how PC impacts model training and contributes to better generalization:

\paragraph{Improved Object Boundary Accuracy.}
In segmentation tasks, one of the most challenging aspects is predicting accurate object boundaries, especially near the edges of objects. By using high-quality pseudo-masks from peer tutors, PC trains the model to better understand where object boundaries are likely to occur. The overlap between the peer object (with a high-quality mask) and the learner object (which might have a low-quality or noisy mask) helps the model refine its understanding of object contours and boundaries.
This improvement in boundary accuracy is critical for segmentation tasks, as even small errors near object edges can significantly degrade performance. With better boundary predictions, the model can more precisely delineate objects, improving overall segmentation quality.

\paragraph{Boosted Instance Discrimination.}
Instance segmentation requires the model to distinguish between different instances of the same class, which can be challenging when objects overlap. PC improves the model's ability to discriminate between instances by forcing it to focus on segmenting specific objects in detail. By copying high-quality pseudo-masks from peer tutors and aligning them with the learner object, PC encourages the model to refine its understanding of individual instances, even when they are similar or overlapping.
This improved instance discrimination is essential for tasks where distinguishing between multiple objects of the same category is critical. By training on augmented data that emphasizes instance separation, PC helps the model avoid misclassifications and improves its accuracy in handling complex scenes with multiple objects.

\paragraph{Reduction of Overfitting.}
By copying high-quality pseudo-masks from peer tutors and pasting them onto different images, PC increases the variability of object combinations and object-background interactions that the model can learn from.
The augmented dataset allows the model to learn more generalized features, such as how different objects interact in various contexts, which helps reduce overfitting to specific patterns in the training data. The model becomes more robust and adaptable to a wider variety of real-world data scenarios.

%\fi

\end{document}